\documentclass[manuscript,screen,dvipsnames]{acmart}
\usepackage{graphicx}
\usepackage{subfig}
\usepackage{tikz}
\usetikzlibrary{positioning}
\usepackage{dsfont}
\usepackage{pifont}
\usepackage{xcolor}
\usepackage{booktabs}
\AtBeginDocument{%
  \providecommand\BibTeX{{%
    \normalfont B\kern-0.5em{\scshape i\kern-0.25em b}\kern-0.8em\TeX}}}
\newcommand{\argmax}[1]{
    \underset{#1}{\mathrm{argmax}}\
}
\usepackage{hyperref}
\hypersetup{
    colorlinks=true,
    linkcolor=blue,
    filecolor=magenta,      
    urlcolor=cyan,
    pdftitle={Overleaf Example},
    pdfpagemode=FullScreen,
    }
\usepackage{colortbl}
\newcommand{\Gu}{\textsc{G1}}
\newcommand{\Gd}{\textsc{G2}}
\newcommand{\Gt}{\textsc{G3}}
\newcommand{\Gq}{\textsc{G4}}
\newcommand{\Gc}{\textsc{G5}}
\newcommand{\Gs}{\textsc{G6}}

\newcommand{\polblogs}{\textsc{Polblogs}}
\newcommand{\lastfm}{\textsc{LastFM}}
\newcommand{\fbe}{\textsc{Facebook-E}}
\newcommand{\fbp}{\textsc{Facebook-P}}
\newcommand{\pokec}{\textsc{Pokec}}
\newcommand{\dblp}{\textsc{DBLP}}
\newcommand{\movielens}{\textsc{MovieLens}}
\newcommand{\nba}{\textsc{NBA}}
\newcommand{\google}{\textsc{Google+}}
\newcommand{\reddit}{\textsc{Reddit}}

\newcommand{\no}{\textcolor{red}{\ding{55}}}
\newcommand{\yes}{\textcolor{ForestGreen}{\ding{51}}}

\definecolor{blue}{rgb}{0.94, 0.97, 1.0}
\definecolor{cosmiclatte}{rgb}{1.0, 0.97, 0.91}
\definecolor{gainsboro}{rgb}{0.86, 0.86, 0.86}
\definecolor{mistyrose}{rgb}{1.0, 0.89, 0.88}
\setcopyright{acmcopyright}
\copyrightyear{}
\acmYear{}
\acmDOI{}

\acmBooktitle{}
\acmPrice{}
\acmISBN{}

\newcommand{\charlotte}[1]{\textcolor{black}{#1}}
\newcommand{\cris}[1]{\textcolor{black}{#1}}
\begin{document}

\title{A Survey on Fairness for Machine Learning on Graphs}

\author{Charlotte Laclau}
\affiliation{%
  \institution{Télécom Paris, Institut Polytechnique de Paris, }
  \streetaddress{}
  \city{}
  \country{France}
}

\author{Christine Largeron, Manvi Choudhary}
\affiliation{%
  \institution{Univ Lyon, UJM-Saint-Etienne, CNRS -- Institut d Optique Graduate School, Laboratoire Hubert Curien UMR 5516}
  \streetaddress{}
  \city{Saint-Etienne}
  \country{France}
}


\renewcommand{\shortauthors}{}

\begin{abstract}
Nowadays, the analysis of complex phenomena modeled by graphs plays a crucial role in many real-world application domains where decisions can have a strong societal impact. However, numerous studies and papers have recently revealed that machine learning models could lead to potential disparate treatment between individuals and unfair outcomes. In that context, algorithmic contributions for \cris{graph machine learning} are not spared by the problem of fairness and present some specific challenges related to the intrinsic nature of graphs: (1) graph data is non-IID, and this assumption may invalidate many existing studies in fair machine learning, (2) suited metric definitions to assess the different types of fairness with relational data and (3) algorithmic challenge on the difficulty of finding a good trade-off between model accuracy and fairness. This survey is
dedicated to fairness for relational data. It aims to present a comprehensive review of state-of-the-art techniques in fairness on \cris{graph learning} and identify the open challenges and future trends. In particular, we start by \cris{ positioning our paper in relation to previous works and by specifying its scope with a focus on two classical graph learning tasks: edge prediction and node classification. We motivate research in this field by reporting the legal framework relating to the use of Artificial Intelligence (AI) and by providing examples of  sensible applications.} 
We recall the different metrics proposed to evaluate potential bias at different levels of the graph \cris{learning} process; then we provide a comprehensive overview of recent contributions in the domain of fair machine learning for graphs, that we classify into pre-processing, in-processing, and post-processing models. We also propose to describe existing graph data and synthetic and real-world benchmarks. Finally, we present in detail five potential promising directions to advance research in studying algorithmic fairness on graphs. We hope that this survey will motivate researchers to tackle these issues in the near future and believe that it could be attractive to researchers and practitioners in areas including machine learning, artificial intelligence, and social science.  Additional materials, including codes and datasets, to complement this survey are available at: \url{https://github.com/laclauc/fairness-graph/}.
\end{abstract}


\maketitle

\section{Introduction}
We live in a world where an increasing number of decisions, with major societal consequences, are made or at least supported by algorithms that diligently learn the patterns from a training sample and gain their discriminating ability by identifying the key attributes correlated with the desired output. These attributes, however, can represent sensitive information that, in turn, can lead to a significant bias in the model’s predictions when deployed on a previously unseen
sample. In this context, an important issue that arises is whether the decisions made or supported by such algorithms are fair. 

\smallskip

Several competing and contrasting definitions of what it means to be \textit{fair} for a Machine Learning (ML) model have been proposed in recent literature. One can broadly categorize these definitions into three classes: (1) group fairness, which emphasizes disparate treatment of individuals with respect to some sensitive attribute defining groups. The underlying idea of group fairness is that minority groups should receive similar treatment as that of advantaged groups \citep{ustun2019fairness, hardt2016}; (2) individual fairness, which requires that similar
individuals should be treated similarly \citep{DworkHPRZ12}; 
and finally (3) counterfactual fairness, where the underlying idea is that a fair decision for a given individual should not depend on the change in the value of his/her sensitive attribute \citep{kusner2017}. Selecting the appropriate metric is intrinsically task-dependent and one of the challenges remains in the fact that the literature has mostly focused on tabular data so far, leaving more complex data slightly aside. In the subsequent sections, we intend to give an overview of the existing metrics adapted or dedicated to graphs and highlight their current weaknesses. 

\smallskip

The problem of biased decision (i.e. unfair situation) in ML can stem from different sources along the decision-making process: the sample data from which the algorithm is learning, the machine learning model used and the exploitation of the output obtained with these models. As a result, contributions in the field of algorithmic fairness can be roughly divided into three categories: the pre-processing methods, whose goal is to remove the bias from the original data itself; the in-processing methods in which the problem of fairness is addressed at the learning time usually by including fairness constraints in the objective function; and the post-processing methods, which aims to remove the bias from the output of the algorithm used to solve the task at hand. 
These methods are based on various techniques, including but not limited to adversarial learning \cite{adel2019one,madras2018learning,xu2019achieving}, causal methods \cite{nabi2018fair,glymour2019measuring,wu2019counterfactual}, optimization under constraints \cite{kim2018fairness,celis2019classification,cotter2019optimization} and sampling methods \cite{oneto2019taking,dwork2018decoupled}. They have been explored in detail in recent surveys \cite{caton2020fairness, barocas2016big, mehrabi2021} and we refer the interested reader to them for more details but we provide a description of these surveys in Section \ref{subsec:related}.
However, while these solutions have proven efficient in mitigating potential bias, they were all designed for tabular data (vectors). In this survey, we are interested in fairness for graph data, or fair graph learning, a recent but fast-growing research area.

Graphs present several specificities making it difficult to use existing fair approaches, originally developed for standard tabular data. Indeed, graphs are non-iid (independent and identically distributed) and non-euclidean data. The first assumption implies that changing or altering information about a given node (attributes or connections) in the graph will impact its neighbors. As a result, evaluating and mitigating potential bias require proper handling of this assumption. \cris{With the development of embedding models, a} second assumption implies that before learning most of the \cris{recent} models (classification, ranking, or clustering), one should first learn a representation of the graph in the form of vectors. At the node level, this corresponds to node embedding models. And, as we will demonstrate in this survey, depending on the objective function used in these models, the learned node representation can capture different amounts of bias, or make it more difficult to mitigate the bias afterward. This last challenge is closely related to the more general problem of fair representation.

\paragraph{Fair Representation Learning} 
Inducing fairness in machine learning led to the focus on learning fair representations for complex data. The first paper to address this problem \cite{zemel2013learning} proposes an algorithm for fair classification which is capable of both group and individual fairness and, that aims to encode the original information from tabular data while at the same time ignoring the information of the protected group. In the same spirit, adversarial regularization techniques have been proposed to straightforwardly learn fair representations as done by \cite{madras2018learning} and \cite{edwards2016}. Similarly, in \cite{creager2019flexibly}, authors propose to learn a representation that achieves group and sub-group fairness.  Their representations are modifiable at test time both simply and compositionally to achieve sub-group fairness with respect to multiple protected attributes. With a similar goal to learn an invariant representation, in \cite{louizos2016variational} the authors propose a semi-supervised fair model based on the Variational Autoencoder \cite{kingma2013auto,rezende2014stochastic} and further guarantee invariance by using a Maximum Mean Discrepancy based regularization  \cite{NIPS2006_e9fb2eda}. Insofar as to deal with graphs, it is essential to start by learning a representation in a Euclidean space, most of the existing contributions in fair graph mining belong to this family of approaches. These methods are referred to as fair node embeddings and we describe them further in Section \ref{sec:fair_embedding}. 

\medskip 

\paragraph{Research Questions} In this survey, we aim to address the three following fundamental research questions about fairness on graphs : 
\begin{itemize}
    \item[\textbf{Q1}] \textbf{Can we tackle bias by altering the graph structure itself?} To answer this question, we provide an overview of existing techniques that have been proposed to reduce bias by transforming the original graph. 
    \item[\textbf{Q2}] \textbf{How much of the bias is present in node embedding?} Here we are interested in understanding to what extent some commonly used node embedding techniques transcribe potential bias. This question relates to how to control the bias at training time. We \cris{notably} present node embedding methods that include fairness constraints at the learning stage. \cris{We have chosen to distinguish on the one hand unsupervised node
embedding approaches, that build upon existing node embedding models and, on the other hand, models which are solving the problem
in an end-to-end fashion by specifying the type of task considered.}
    \item[\textbf{Q3}] \textbf{Is the model outcome fair?} This last question is linked to the post-processing family of models existing for tabular data, as when reaching this step, we are usually dealing with vectors. We will present existing models that propose to mitigate bias at the task level for graphs.
\end{itemize}

To cover these questions in a thorough manner, we organize our article as follows. 
After situating our contribution and specifying its scope, we start by introducing the problem of fairness for graph-related tasks including node embedding, node classification, and edge prediction in Section \ref{sec:setup}. Then, we recall the regulations in force in AI related to fairness and, we also present two real-world application cases where unfair graph learning algorithms can lead to discriminatory outcomes as well as specific fairness challenges related to graph data in Section \ref{subsec:application}.
Then, we proceed with a presentation of the existing metrics used to measure and evaluate bias in Section \ref{sec:metrics}, and differentiate based on the level of the graph on which they operate.  Sections \ref{sec:preproc}, \ref{sec:fair_embedding} and \ref{post-processing} present the main algorithmic contributions for fair graph learning. These sections cover  \cris{pre-processing methods (Section  \ref{sec:preproc}) whose goal is to remove the bias from the original data itself;
the in-processing methods in which the problem of fairness is addressed at the learning time usually by including
fairness constraints in the objective function  (Section \ref{sec:fair_embedding}) ; and the post-processing methods (Section \ref{post-processing}) which aims to remove the bias from the
output of the algorithm used to solve the task at hand}. Then, in Section \ref{sec:data} we provide an overview of both synthetic and real-world benchmarks that are commonly used to evaluate contributions in fair graph mining. Finally, we conclude this article by discussing some important remaining open challenges in Section \ref{sec:conclusion}.

\section{Related Work, Methodology and General Set-up}\label{sec:setup}
In this section, we give an overview of existing surveys and tutorials on the topic of fairness in machine learning and graphs. We also define the general framework of fair graph mining and motivate the importance of this research field with two concrete examples.

\subsection{Related Surveys and Tutorials}\label{subsec:related}
\label{Surveys}

Before dealing with the fairness problem for graph-type data, the subject was first approached for tabular data and there exist already several surveys  on the broader topic of fairness. We only mention a few below.
The first one is an online book \cite{barocas2019}, where the authors give a very broad overview of the challenges posed by fairness in machine learning. In addition to the more traditional chapters on existing approaches and metrics, they notably provide discussions on the societal aspects of the problem. For instance, on the importance and difficulty of testing discrimination of in-production machine learning systems or the relations between machine learning structural, organizational, and interpersonal discrimination in society. In a similar spirit, in \cite{beutel2019putting}, the authors develop a case study on the application of fairness in machine learning research to a classification system in production. This type of study, while not reviewing all existing approaches in the domain, offers interesting new insights into how we can test, measure and address algorithmic fairness in practice. In \cite{mehrabi2021}, the authors proposed also an interesting covering of existing algorithmic methods to achieve fairness in machine learning. 

\cris{Complementing these previous general surveys, other papers question the link between the measures proposed in the literature (Demographic parity, Equalized Odds, \textit{etc}) and the type of equity apprehended. For instance, in  \cite{Mitchell2021}, authors highlight the gap between the formal definitions and the objectives in terms of fairness which are supposed to be addressed, while in \cite{Tang2023} authors revisit the notion of equity from a causality perspective.
There are also more targeted surveys such as \cite{wan2022} which presents a typology of in-processing techniques proposed in the literature to mitigate bias during the learning process or \cite{DuYZH21} that focuses on fairness in deep learning with a computational perspective and concludes that for this type of model, interpretability can help to better understand the reason that leads to unfair algorithms.
Finally, we can mention \cite{LeQuy2022} who list datasets used for fairness-aware machine learning evaluation even if it only concerns tabular data as it is the most common representation in ML but, in this sequel, we describe several real graph datasets and, we propose a generative process to build synthetic graphs for fairness-aware edge prediction evaluation.}

As said previously, we want to emphasize that the aforementioned references are not dedicated to graph data, contrary to our work, whose structure and objectives resemble the most to \cite{mehrabi2021} but with a focus on fairness for machine learning on graphs. Indeed, it is only more recently that the problem of fairness in graphs has attracted the interest of the community as evidenced by the organization of tutorials in conferences in the field such \textit{Fairness in Networks} \cite{Venkatasubramanian2021} at KDD 2021,  \textit{Fair Graph Mining} \citep{kang2021} at CIKM 2021 or \textit{Fairness on Graphs: State-of-the-Art and Open Challenges} \cite{Kang2022} at KDD 2022. This has made it possible to better understand and define the notion of fairness in this context, to propose measures to assess it, often drawing inspiration from those already designed for tabular data, and to develop learning algorithms aimed at treating more fairly classical tasks such as the supervised or unsupervised classification of nodes of a graph, prediction of links and recommendation or influence maximization problem. However,  to our knowledge, there are still very few published papers offering a general overview of the issue of fairness in the processing of graph data except  \cite{dong2023, Pitoura2021}. The first one \cite{dong2023} is dedicated to graph data mining algorithms and the second \cite{Pitoura2021} deals with fairness for rankings and recommendations  whereas this survey considers mainly fairness in graph-based machine learning  and provides a comprehensive and organized review of the methods according to the stage where fairness is imposed during the ML process: the pre-processing based on a transformation of the input to  repair the graph at the origin,  with in-processing approaches by adapting or designing fair algorithms or, during the post-processing by removing the bias from the output of the algorithm. 
Moreover, we restrict ourselves to group, individual, and counterfactual fairness, as they are the most common forms of fairness studied in this context. 

\subsection{Scope of the Survey}

For this survey, we select papers based on the following criteria:
\begin{itemize}
    \item Papers published before March 2023. Note that the first contributions date from 2019. 
    \item Papers published in proceedings of conferences or in a journal. Note that we choose to exclude papers only available on Arxiv and short papers presented in workshops without associated proceedings. 
    \item We only select papers that consider input data that are naturally structured as graphs. Indeed, we chose to exclude papers that were using graphs as a tool to inject fairness, a common approach in Recommender Systems (RS). 
\end{itemize}
Overall, based on these criteria, we selected 30 contributions, that are summarized in Table \ref{tab:sum_methods} and will be further described in this survey. 

\subsection{Problem Setup}\label{subsec:setup}
There exist multiple tasks that we can address when dealing with complex data represented by graphs. We can divide these tasks according to the level of the graph on which they operate: we distinguish graph-level (e.g. predict the property of an entire graph), from node-level (e.g. predict the role of a node or detect communities of nodes within a graph) and edge-level (e.g. predict the existence of an edge) tasks. For more details about these various tasks, we refer the interested reader to \cite{AggarwalWang2010, xia2021}. In this survey, we will focus on node classification and edge prediction. One example of edge-level inference is, given a pair of nodes, to predict the relationship between them. We can phrase this as an edge-level classification: given a pair of nodes, we wish to predict which of these nodes share an edge or what the value of that edge is. For instance, on a social platform, edge prediction amounts to predicting potential friendship relations between users, while node classification amounts to predicting a label for each user, that encodes interests or hobbies. In addition to this label, in the context of fairness, we also assume that nodes have one or more attributes, some of which are considered protected. Thus in the above example, users of the social platform might have entered their demographic information, e.g. their gender. In this context, we wish to ensure that edge prediction models or node classification ones will not depend on this factor. 

\begin{figure}[!t]
    \centering
    \includegraphics[width=\textwidth]{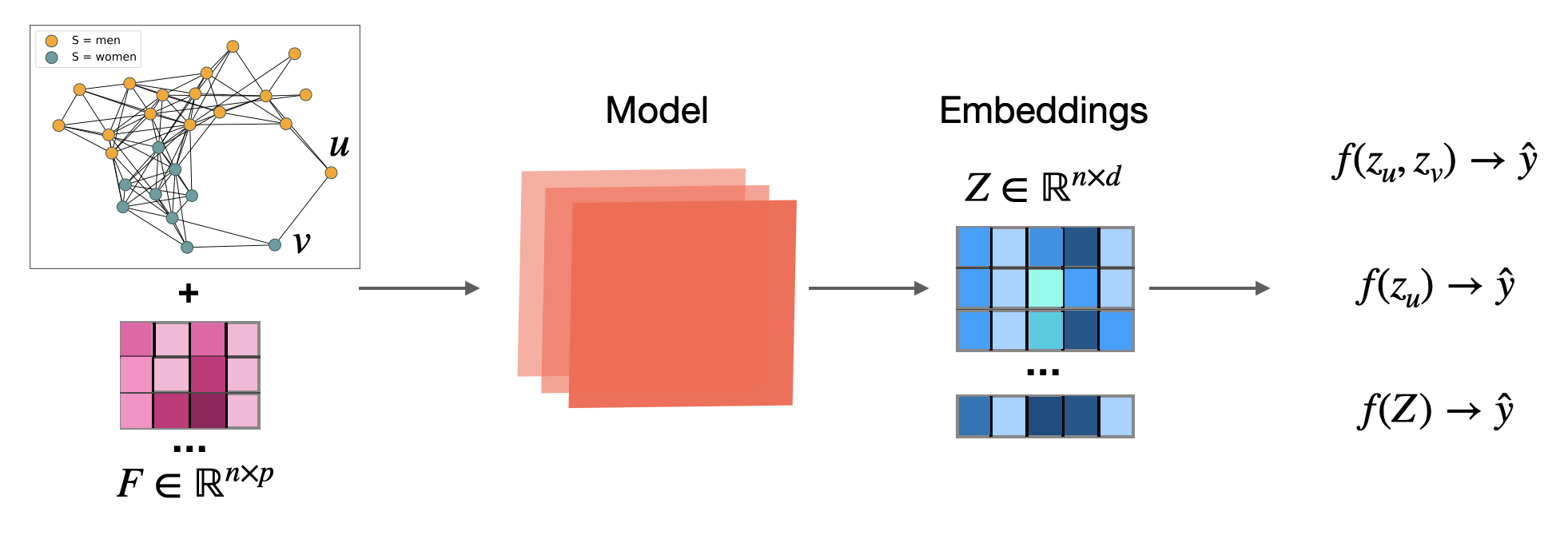}
    \caption{Machine Learning on graphs. From the original graph (left) and eventually additional feature information (including the protected attribute), a mapping function (model) is learned that produces the matrix of node embeddings $Z$ (middle). These embeddings are then used to solve several types of graph problems (right): edge prediction (top),  node classification (middle), or graph classification (bottom).}
    \label{fig:process}
\end{figure}

\paragraph{Notations} Throughout this article, we will use the following notations. 
Let $\mathbb{V}$ denotes an abstract vertex space, and let $\mathcal{V} = \{v_1, \dots, v_{n}\} \in \mathbb{V}^n$ be a set of $n$ vertices drawn from an arbitrary  distribution over $\mathbb{V}$. In the finite case, we further consider a graph $\mathcal{G}=(\mathcal{V}, \mathcal{E})$
where $\mathcal{E} \subseteq \mathcal{V}\times \mathcal{V}$ defines the set of edges in the graph. \charlotte{$\mathcal{G}$ can also be represented through an $n$-by-$n$ adjacency matrix $X$, where $X$ encapsulates the structural information of
the graph, in which a non-zero entry indicates there is a connection between two nodes $v_i$ and $v_j$ (e.g. if $x_{i,j}=1$)}. Furthermore, we can consider different labels. For edge prediction, \cris{the label  characterizes the presence or absence of a link between two nodes.} Thus, we consider $y$ such that a tuple $\{(v_i, v_j, y(v_i,v_j)\}$ defines the existence (i.e., $y(v_i,v_j)=1$) or the absence (i.e., $y(v_i,v_j)=0$) of an edge $(v_i,v_j) \in \mathcal{E}$ and $m$ is the number of edges in $\mathcal{E}$.
\cris{For node classification, the label relates to the node and not the link and it makes it possible to characterize the class $y(v_i) \in Y$ of the node $v_i$.} Furthermore, we assume the existence of a variable that explicitly encodes some sensitive information and use the letter $A$ to designate a discrete random variable that captures one or multiple sensitive characteristics. For instance, $A_{v_i}=0$ means that the node $v_i$ is assigned with the sensitive group labeled as $0$. \cris{In the sequel, k denotes the number of modalities of $A$, or equivalently the number of sensitive groups, and $|A| > 1 $ means that there are several sensitive characteristics}. Note that, one of the main challenges of dealing with such settings, is that the sensitive information might as well be implicitly encoded in the graph structure, requiring to consider this latter when trying to achieve fairness. It can be notably the case in the presence of homophily when nodes sharing the same attribute value are more likely to be connected. 

\paragraph{Node embedding} As machine learning models typically take tabular data (vectors) as input, one needs to add an intermediate step between the original graph representation and the task to be solved to apply these models with the graph data. Node embedding consists in learning a (non-)linear mapping of each node $v\in \mathcal{V}$  into a vector space $f: \mathcal{V} \rightarrow \mathbb{R}^d$ where $d$ is the number of dimensions of the vector space. We further assume that $d\ll n$. These vectorial representations of the nodes are stored in a matrix $\textit{Z} \in \mathbb{R}^{n \times d}$ and referred to as node embeddings.
Once learned, these node embeddings can be used for any downstream tasks, including edge prediction, community detection, or node classification to name a few, as illustrated in Figure \ref{fig:process}. In the context of fair machine learning, we further expect that these learned node embeddings do not reflect potential bias arising from the original graph, and are captured by the sensitive attribute $A$. To sum up, node embeddings should have the following properties.
\begin{enumerate}
\item[\textbf{P1.}] Embeddings should reflect different properties of the graph structure, e.g., two nodes that are close in the graph should also be closed in the embedding space. \charlotte{More precisely, from the individual fairness perspective, one will expect that two nodes having similar properties in the graph should have similar embeddings.}
\item[\textbf{P2.}] Embeddings should be independent of one or multiple sensitive attributes, hence one should not be able to retrieve information regarding sensitive attributes from these embeddings. This property refers to group fairness. 
\end{enumerate}
One should note that having \textbf{P1} and \textbf{P2} is rather a difficult task, especially when the structure or \charlotte{other existing node features} of the graph are strongly correlated with the protected attribute(s). Consequently, the objective here is usually to find a reasonable trade-off between these properties. Furthermore, while \textbf{P1} is directly ensured by the objective function used to learn the embeddings, \textbf{P2} can be obtained in different manners : (1) by pre-processing the original graph to remove potential bias in the data structure before learning the embeddings; (2) by making change, e.g. adding constraints, to the objective function used to learn the embedding; (3) by post-processing the obtained node embeddings to filter out the bias. 

\paragraph{Machine Learning Tasks on Graphs} We now describe the three most common tasks, namely node classification, edge prediction, and community detection. On the one hand, in graphs, the nodes are often characterized by contextual information in the form of node attributes. When the values of these attributes (or labels) of some nodes are only partially known, node classification or semi-supervised label prediction can be used to determine these latter \cite{bhagat2011node}. Formally, the goal is to learn a function $h:\mathbb{V}\rightarrow Y$, where $Y$ is a label set, such that $h$ is as close as possible to $y$, the true node class. On the other hand, edge prediction considers tuples of nodes as input. Given a pair of nodes, edge prediction consists in determining if an edge exists between them or not. Formally, it aims at finding a function $h:\mathbb{V}\times \mathbb{V} \rightarrow \{0,1\}$ such that $h$ is as close as possible to $y$, the true edge label.
These two tasks are supervised, hence the notion of fairness for them is usually defined in terms of the predictions made by the function $h$ conditionally on the protected attributes. As we will see in the next section, different definitions of fairness for supervised tasks have been proposed, and they all capture different aspects of the problem. 
Finally, community detection consists in discovering cohesive groups or clusters in a complex network. Most of the time, this is done in such a way that nodes belonging to the same community are densely connected to each other while sparsely connected to the rest of the network. This task is unsupervised and for further details, we refer the interested reader to \cite{fortunato2010community, Su2021}, but we can notice that the community detection methods  are often based on properties of the nodes and the graph, such as assortativity or homophily of the graph, that, if exploited can result in more bias (see Section \ref{sec:metrics}). 
As contributions to fair community detection are more scarce, in the sequel, we  focus mainly on fair node classification and fair edge prediction in the remainder of this paper.

\section{Motivation for Research on Fairness for Graphs}\label{subsec:application}
In this section, we present the legal framework surrounding the use of artificial intelligence. It should be noted that this framework is not specific to graph data, but encompasses the latter. Based on this legal framework, we then describe three applications identified as high-risk and for which the input data is structured in the form of graphs. We conclude this section by presenting the scientific challenges of solving fairness for graphs.
\subsection{Legal Background}
We are now ready to detail the existing regulation of the usage of AI. For this survey, we focus on the fairness dimension and chose to leave other, equally important, concepts including privacy and transparency to name a few.  We choose to focus on the USA and the European Union.
\paragraph{European AI Act}
The European AI Act is a proposed regulation by the European Union that aims to provide a legal framework for the development and use of artificial intelligence (AI) in Europe. The proposed regulation covers a wide range of AI applications, including machine learning. According to the Act, a system is considered fair when it is designed and deployed in a way that respects fundamental rights, prohibits discrimination, and ensures that the interests of all stakeholders are taken into account.
More specifically, the Act defines fairness in the following ways:
\begin{itemize}
    \item Respect for fundamental rights: AI systems should respect the fundamental rights and freedoms of individuals, including privacy, non-discrimination, and the right to a fair trial.
    \item Prohibition of discrimination: AI systems should not discriminate against individuals based on their personal characteristics such as race, gender, age, disability, or sexual orientation.
    \item Transparency and explainability: AI systems should be transparent and explainable, meaning that their decision-making processes and criteria should be clear and understandable to both users and affected individuals.
\end{itemize}
In this survey, we focus on fairness defined as an absence of discrimination against individuals on the basis of sensitive characteristics.

\paragraph{USA Regulation of Machine Learning}
In the US, there are several legal obligations related to fairness, particularly in the context of employment and lending practices.
One of the most important legal obligations related to fairness is contained in Title VII of the Civil Rights Act of 1964. This law prohibits discrimination in employment on the basis of race, color, religion, sex, or national origin. 
Another important law related to fairness is the Equal Credit Opportunity Act (ECOA), which prohibits discrimination in lending on the basis of race, color, religion, national origin, sex, marital status, age, or receipt of public assistance. Lenders are required to ensure that their lending practices are fair and do not discriminate against individuals based on these protected characteristics.

Overall, the legal obligations related to fairness in the US are designed to ensure that individuals are treated equally and that discrimination is not allowed in employment, lending, or other contexts.

However, as indicated in Section \ref{Surveys}, it should be noted that previous work \cite{Mitchell2021, Tang2023} questioned the ability of metrics and models introduced in AI to understand the notion of discrimination due to its complexity, but this is not the purpose of this paper.

\subsection{High-risk Applications and Challenges}
Hereafter, we provide important societal examples of applications where the problem of fairness when dealing with graphs can occur. Note that, this is obviously not an exhaustive list of applications, but these examples motivate us to believe that working on that particular topic is of high importance. We will also use them in the remainder of the paper to illustrate some important concepts or intuitions.

\begin{figure}[t]
    \centering
    \subfloat[Job recommendation]{ \includegraphics[width=0.33\linewidth]{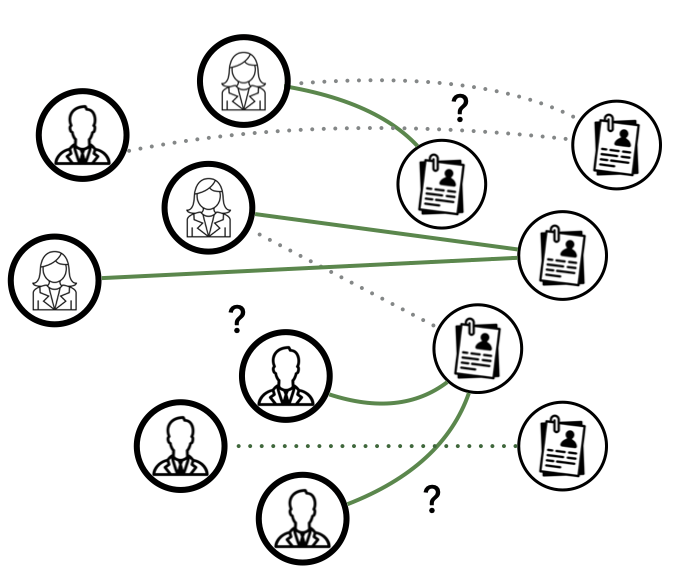}} 
    \subfloat[Online polarization]{
    \includegraphics[width=0.33\linewidth]{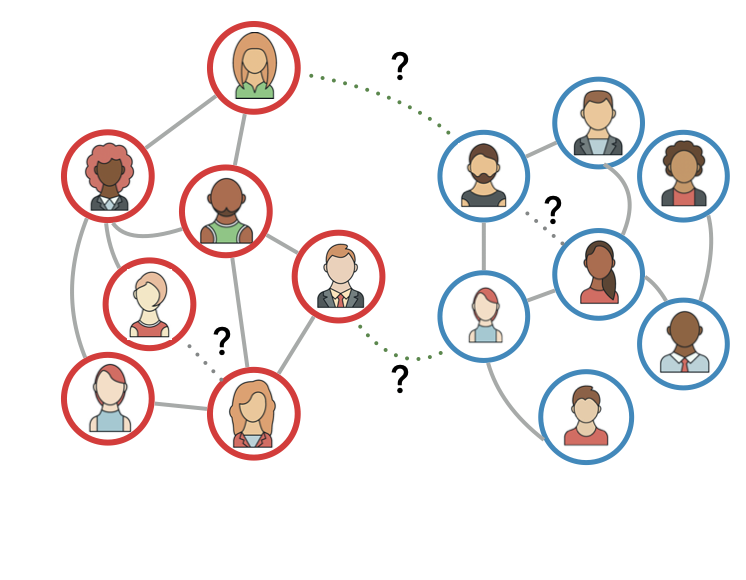}}
     \subfloat[Role of influencers]{
    \includegraphics[width=0.33\linewidth]{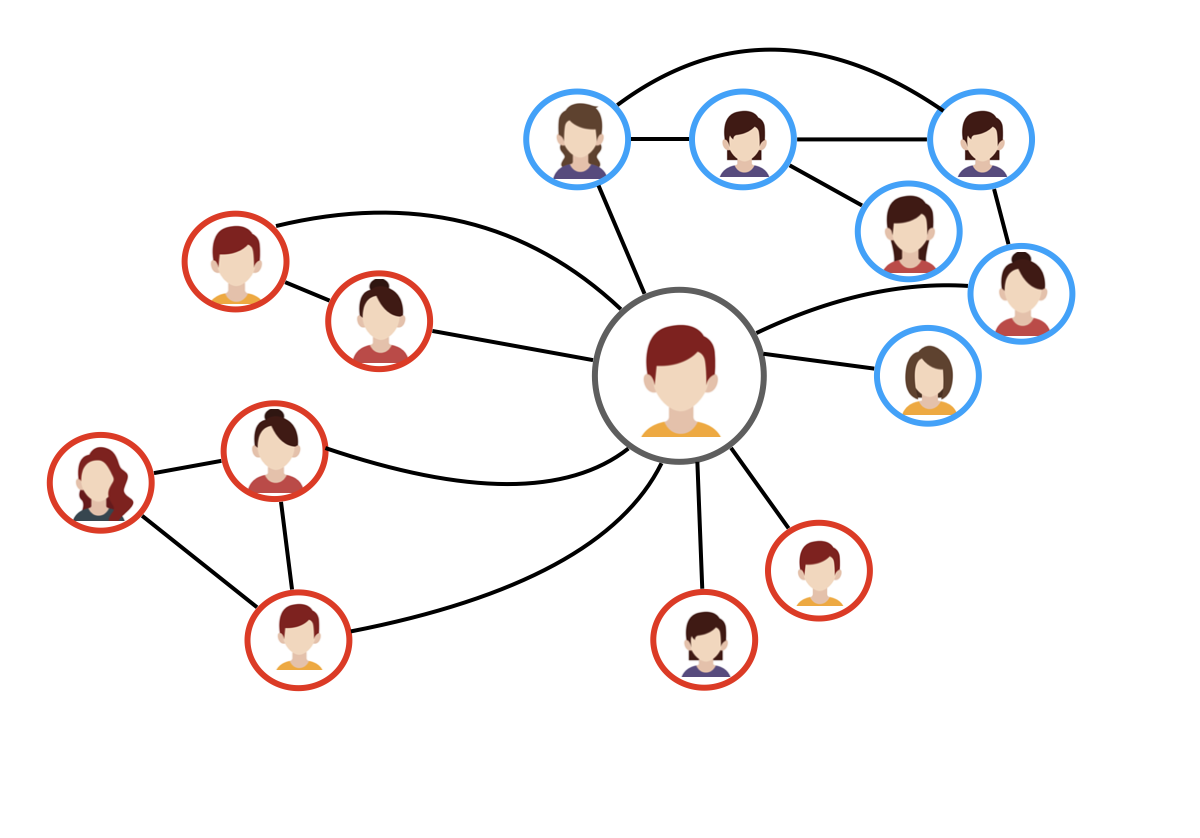}}
    \caption{Examples of networks with two different potential bias: job-candidate matching (a) and a social media network with political echo chambers (b) and in the presence of an \textit{influencer} node (c).}
    \label{fig:example1}
\end{figure}

\paragraph{Social Networks}
For our first example, we consider a social network of bloggers (see Figure \ref{fig:example1}(b)): in this context, 
a node represents a blogger, and two nodes are connected if the bloggers follow each other. We also assume that we have access to multiple attributes for each node, including the political inclination and the topics of interest. Political inclination can be considered a protected attribute and such networks present a risk of partisan or political bias. 
Indeed, an edge prediction model might tend to promote only links between people having the same political ideology, since they are more likely to be connected in the network, which leads to the formation of online bubbles and online polarization. Another important source of bias is that social media networks tend to be centralized: a small number of nodes, so-called \textit{influencers}, are at the center of the graphs, meaning that these nodes are connected to nearly all the other nodes (see Figure \ref{fig:example1}(c)). As a result, these \textit{influencers} strongly impact the other nodes around and therefore the outcomes of any supervised or unsupervised models. This example refers to the topic of fair influence maximization which aims to determine a subset of  users  (seed nodes), of a given size $k$,  that maximizes the expected number of influenced users in the graph \cite{Kempe2003}.

\paragraph{Professional Network}
The job market and more precisely the job recruitment process is more and more reviewed and supervised using ML algorithms \citep{kenthapadi2017}. As a result, one can ask if these automatic decisions are \textit{fair} for all the applicants (see Figure \ref{fig:example1}(a)). Numerous studies demonstrate the presence of demographic biases in the different aspects of the job market, notably in the recruitment process \citep{Oreopoulos2011,NBERw9873}. In the context of machine learning, a recent paper studies the job platform XING, a professional social network, and shows that it exhibits discrimination with respect to gender, by ranking less qualified male candidates higher than more qualified female candidates \citep{Lahoti2019}.
This problem of candidate/job matching can be represented by a bipartite graph with two types of nodes: jobs and applicants. Both types come with attributes. For instance, for the applicants, we might have access to civil information, job title, and previous salary. For the jobs, we can have access to a textual description and the proposed range for the salary. 
In this example, \textit{fair} can have different meanings: on the one hand, we expect that a recommendation (link prediction between an applicant and a job) algorithm is independent of some sensitive attributes of the candidates, including, for instance, their gender or ethnic origin; we might also expect from a node regression model aiming to predict the salary of each candidate to also be independent of such attributes. This type of fairness is referred to as group fairness in the literature. Finally, we also expect the recommendation to remain fair from an individual point of view, i.e., two candidates with similar skills should obtain similar results. This type of fairness is usually called individual fairness in the literature. 

\begin{table}[!ht]
    \centering
     \caption{Summary of all methods reviewed in this paper and a link to the available implementations when available.
     Row colors groups methods falling in the same category.}
    \label{tab:sum_methods}
    \resizebox{\textwidth}{!}{
    \begin{tabular}{lccccccl}
    \toprule
     Method&Reference&Year& Code & Pre-processing & In training & Post-processing&Keywords\\
     \midrule
     \rowcolor{blue}
     \texttt{FairOT} & \cite{laclau2020optimal}&2021&\href{https://github.com/laclauc/FairGraph}{Github}&\ding{51}&&&Optimal transport, Laplacian regularization\\
     \rowcolor{blue}
     \texttt{MaxFair}&\cite{jalali2020}&2020&&\ding{51}&&&\\
     \rowcolor{blue}
     \texttt{FairDrop}& \cite{spinelli2021}&2021&\href{https://github.com/ispamm/FairDrop}{Github}&\ding{51}&&&Edge drop, Homophily\\

     \rowcolor{cosmiclatte}
     \texttt{FairWalk}& \cite{rahman2019fairwalk} &2019&\href{https://github.com/urielsinger/fairwalk}{Github}&&\ding{51}&&Random walk\\
     \rowcolor{cosmiclatte}
     \texttt{CrossWalk}& \cite{khajehnejad2021crosswalk} &2022&\href{https://github.com/ahmadkhajehnejad/CrossWalk}{Github}&&\ding{51}&&Random walk\\
     \rowcolor{cosmiclatte}
     \texttt{DeBayes}&\cite{buyl2020}&2020&\href{https://github.com/aida-ugent/DeBayes}{Github}&& \ding{51}&&Conditional network embeddings, Bayeisan prior\\
     \rowcolor{cosmiclatte}
     \texttt{MONET} & \cite{palowitch2020monet}&2020&\href{https://github.com/google-research/google-research/tree/master/graph_embedding/monet}{Github}&&\ding{51}&&GNN, Metadata\\
   \rowcolor{cosmiclatte}
     \texttt{NIFTY}&\citep{agarwal2021towards}&2021&\href{https://github.com/chirag126/nifty}{Github}&&\ding{51}&&GNN, Augmented views, Stability\\
    \rowcolor{cosmiclatte}
     \texttt{FairGNN} &\cite{enyan2021}&2021&\href{https://github.com/EnyanDai/FairGNN}{Github}&&\ding{51}&&GNN, Adversarial learning \\
          \rowcolor{cosmiclatte}
      \texttt{FairVGNN}&\cite{wang2022b}&2022&\href{https://github.com/YuWVandy/FairVGNN}{Github}&&\ding{51}&&GNN, Fairness, feature masks, attribute leakage\\
     \rowcolor{cosmiclatte}
     \texttt{DFGNN}&\cite{OnetoND20}&2020&&&\ding{51}&&\\
     \rowcolor{cosmiclatte}
      \texttt{FGNN}&\cite{zhang2022}&2022&&&\ding{51}&& Fair semi-supervised learning, label propagation, GNN\\
     \rowcolor{cosmiclatte}
        \texttt{FairHIN}&\cite{zeng2021}&2021&\href{https://github.com/HKUST-KnowComp/Fair_HIN}{Github}&&\ding{51}&& \\
     \rowcolor{cosmiclatte}
     \texttt{FairEGM}&\cite{CurrentHG022}&2022&&&\ding{51}&&Demographic parity, GCN, Link prediction\\
  \rowcolor{cosmiclatte}
     \texttt{FairAdj} &\cite{li2021}&2021&\href{https://github.com/brandeis-machine-learning/FairAdj}{Github}&&\ding{51}&&GNN\\
     \rowcolor{cosmiclatte}
     \texttt{UGE}& \cite{wang2022}&2022&&&\ding{51}&&Structural generative graph model\\
     \rowcolor{cosmiclatte}
     \texttt{HM-EIICT}&\cite{SaxenaFP22}&2021&&&&&Link prediction, Similarity-based indices, Social
networks\\
     \rowcolor{cosmiclatte}
     \texttt{FIPR}&\cite{buyl2021}&2021&\href{https://github.com/aida-ugent/FIPR/tree/main/src}{Github}&&\ding{51}&&I-Projection regularizer\\
    \texttt{InFoRM} &\cite{kang2020inform}&2020&\href{https://github.com/jiank2/inform}{Github}&\ding{51}&\ding{51}&\ding{51}&\\
\texttt{REFEREE}&\cite{DongWWDL22}&2022&&\ding{51}&&&GNN, Model explanability\\
    \rowcolor{mistyrose}
     \texttt{CFC} & \cite{bose2019compositional} &2019&\href{https://github.com/joeybose/Flexible-Fairness-Constraints}{Github}&&&\ding{51}&Adversarial learning, Compositional filtering\\
     \rowcolor{mistyrose}
     \texttt{FLIP}&\cite{masrour2020}&2020&\href{https://github.com/farzmas/FLIP}{Github}&&&\ding{51}&Adversarial learning, Modularity\\
     \texttt{EDITS}&\cite{DongLJL22}&2022&&&\ding{51}&&Graph neural networks, Algorithmic fairness, Data bias\\
     \texttt{REDRESS}&\cite{dong2021}&2021&\href{https://github.com/yushundong/REDRESS}{Github}&&\ding{51}&& GNN, Individual fairness, Ranking\\ 
   \texttt{FairInf}&  \cite{RahmattalabiJLV21}&2021&&&\ding{51}&& Influence maximization\\
     \texttt{FairCovering}&\cite{RahmattalabiVFR19}&2019&&&\ding{51}&&Influence maximization\\  
     \texttt{FairTCIM}&\cite{AliBCMGS23}&2023&&&\ding{51}&&Influence maximization, Algorithmic fairness, Social networks\\
  \texttt{DivConstraint}&\cite{TsangWRTZ19}&2019&&&\ding{51}&&Influence maximization\\
  \texttt{AdvFairInfluence}&\cite{khajehnejad2021}&2020&&\ding{51}&&&Influence maximization\\
\texttt{FairSC}&\cite{KleindessnerSAM19}&2019&\href{https://github:com/
matthklein/fair spectral clustering}{Github}&&&& Community detection\\
    \bottomrule
    \end{tabular}
    }
\end{table}

\paragraph{Challenges}
Graph data presents unique challenges for machine learning models in terms of fairness. Here are some examples of fairness issues specific to this type of data, and that were identified in the literature. 
\begin{enumerate}
    \item Structural bias: graphs can have inherent structural biases that reflect historical or societal inequalities. For example, a social network may have fewer connections between people of different races, leading to biased predictions about interactions between people of different races. 
    \item Homophily: homophily refers to the tendency of individuals to form bonds with people who are similar to them \cite{McPherson}. This can lead to a lack of diversity in the data, which can result in biased predictions. For example, a hiring graph that contains only connections between people with similar educational backgrounds may lead to biased predictions about the qualifications of job applicants with different educational backgrounds.
    \item Missing data: Incomplete graphs (either at the link or node level) can lead to biased predictions. For example, if a graph only contains information on the relationships between male employees and their supervisors, a machine learning model trained on this data may make biased predictions about the career paths of female employees.
\end{enumerate}
Overall, fairness in machine learning on graphs (but not only) is a complex and multifaceted concept. Solving fairness-related problems often requires careful consideration of the social and historical context of the data, as well as the use of specialized fairness techniques and algorithms. As we will see throughout this survey, proposed contributions sought to specifically tackle one of the above-mentioned problems at different stages of the machine learning pipeline. These contributions also differ depending on the metrics that they rely on to validate their approaches, with a focus on structural bias, representation bias, or prediction bias.

\section{Evaluation Metrics for Fairness}\label{sec:metrics}
In the following section, we describe metrics proposed in the literature which we classify into structural metrics, representation bias metrics, and metrics acting on the final prediction. 

\subsection{Structural Metrics}
The main challenge when dealing with fairness and graphs is that the topology of the graphs is often strongly correlated with the sensitive attribute that we want to be oblivious to. Consequently, graph-level metrics that were not initially meant to measure a level of fairness have been used when dealing with fairness to (1) obtain insights on the type of accuracy-fairness trade-off that one might be facing for a given graph (i.e. how much accuracy can I expect to lose when enforcing fairness?) and (2) guide approaches to correct this topological bias that should lead to node classification or edge prediction models fairer, for instance by rewiring the biased graph or modifying traditional message passing strategies so as to account for this bias. We refer to these metrics as structural metrics in the following.

\begin{figure}[!ht]
    \centering
    \subfloat[G1, $r=0.51$]{\includegraphics[width=0.25\textwidth]{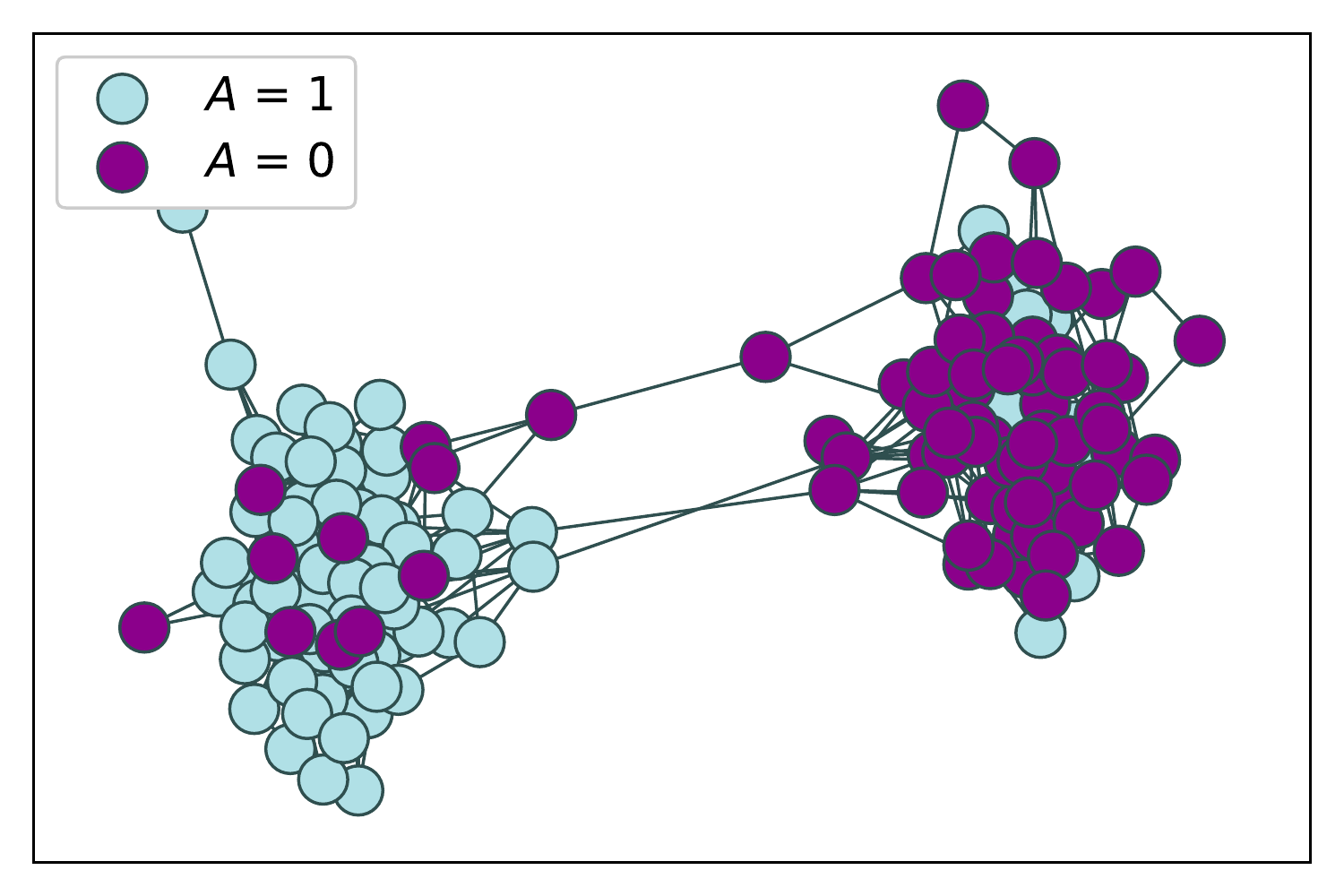}}
    \subfloat[G1 node embeddings]{\includegraphics[width=0.25\textwidth]{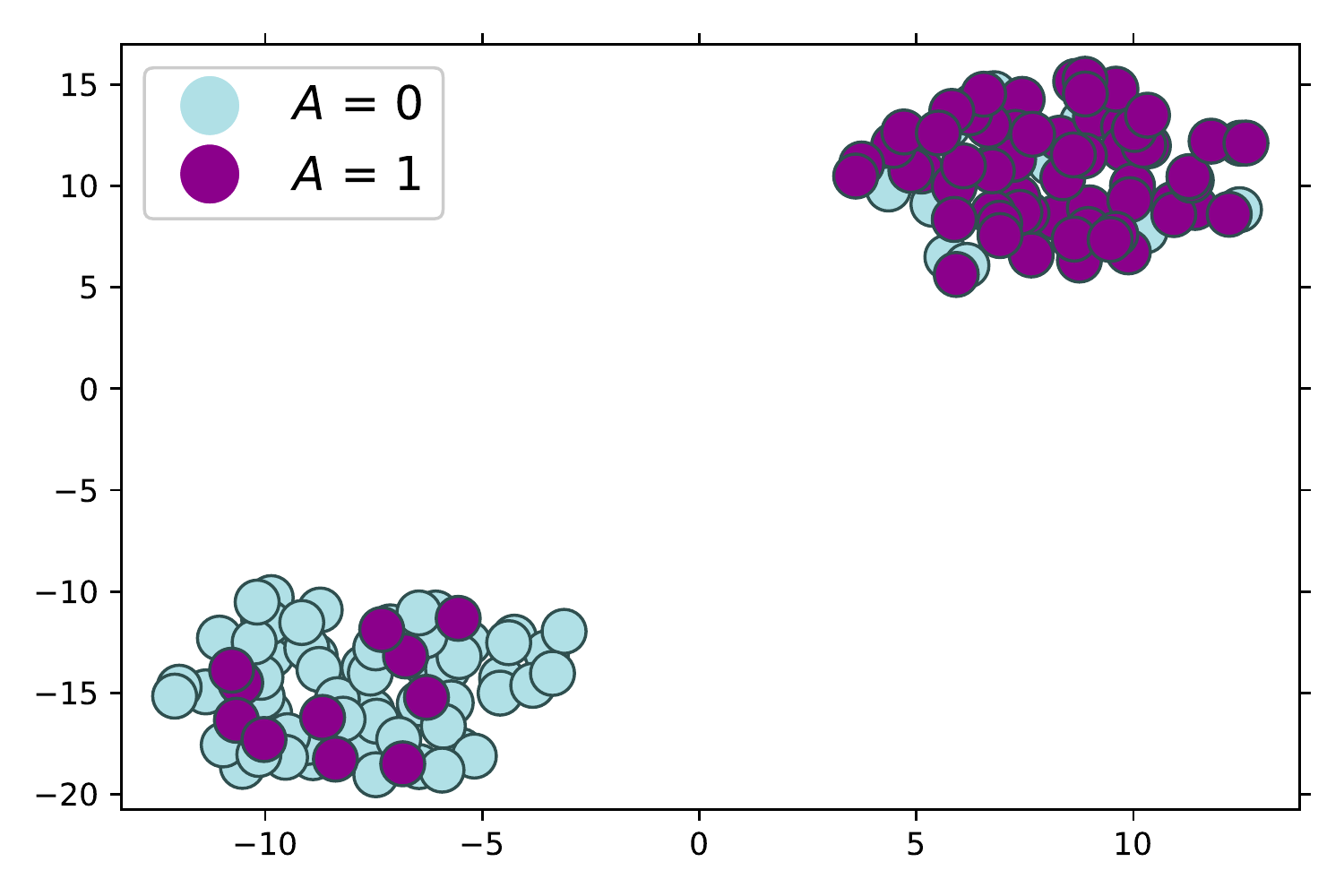}}
    \subfloat[G2, $r=0.04$]{\includegraphics[width=0.25\textwidth]{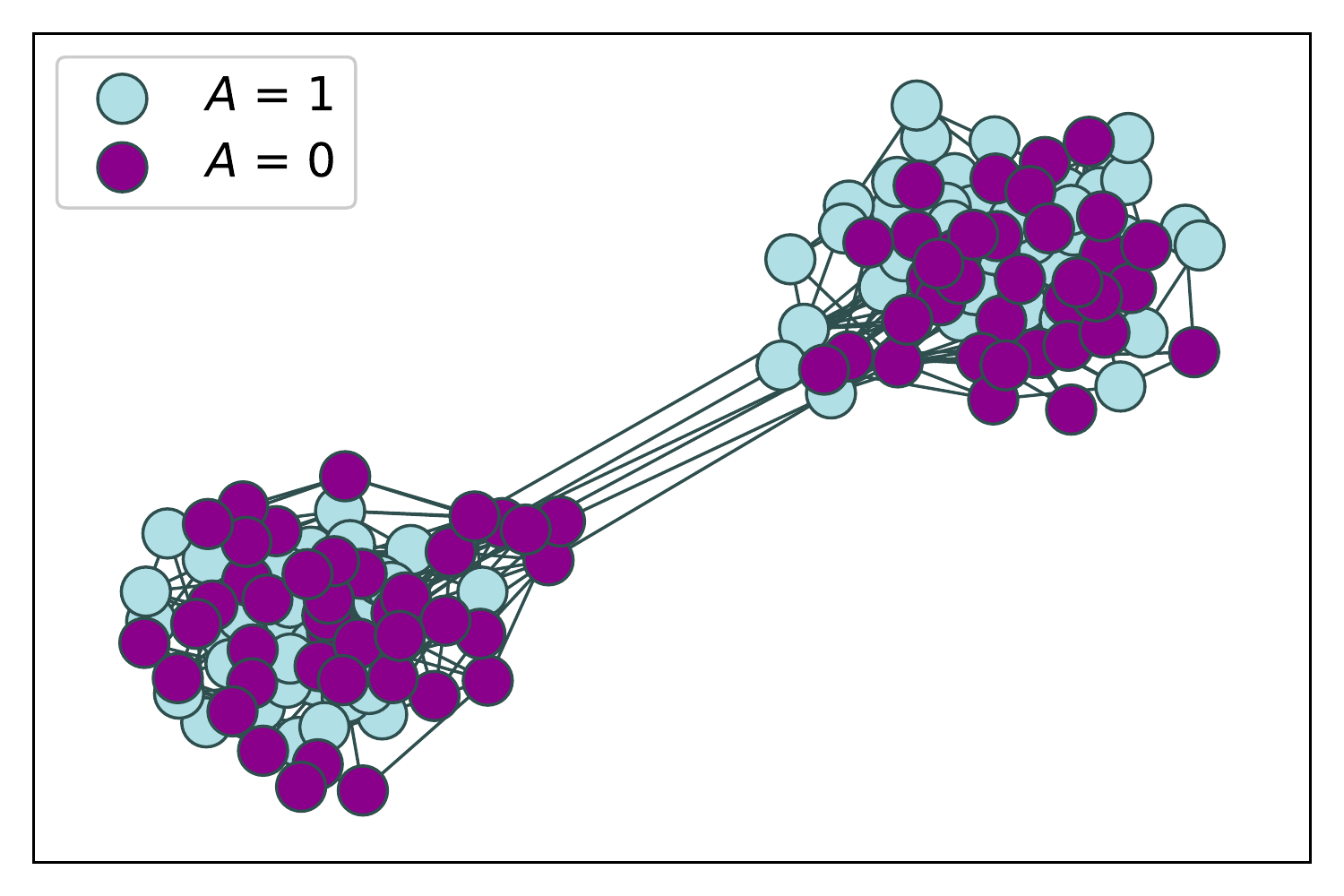}}
    \subfloat[G2 node embeddings]{\includegraphics[width=0.25\textwidth]{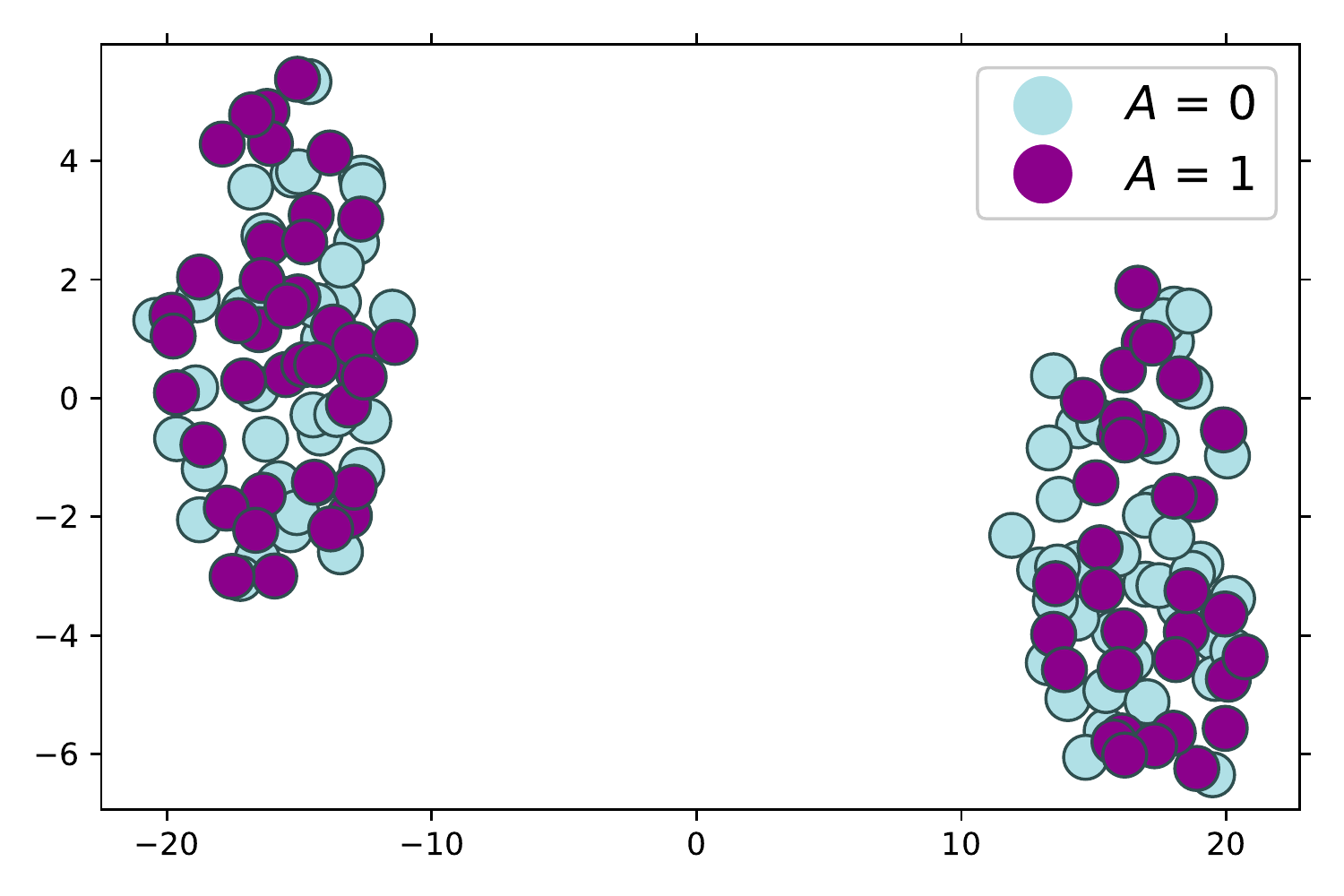}}\\
    \caption{Illustration of the assortative mixing coefficient ($r$). Embeddings are computed with Node2Vec and a dimension of 8. The representation of node embeddings is done with TSNE.}
    \label{fig:assortativity_coefficient}
\end{figure}

Structural metrics aim to capture bias inherent to the graph topology. This aspect is of particular importance for fair graph machine learning. In this context, the notion of homophily of the graph for a particular attribute, which can be evaluated using the Assortative Mixing Coefficient \cite{Newman_2003}, is often used to assess the bias inherent to the graph structure itself. Indeed, mixing in social network analysis studies the tendency of nodes to connect with other nodes having similar attributes. In the context of dyadic fairness, the underlying idea is that the stronger the correlation between the community structure and the protected attribute (i.e. a coefficient with a value close to 1 or -1), the higher the risk of bias along the process. Figure \ref{fig:assortativity_coefficient} illustrates the idea. Thus, the graph $G1$ (a) has two communities, and these communities are almost completely defined by the protected attribute $A$. Going back to our online bubble example, where $A$ is the political inclination of the bloggers with $A=1$ for Republicans and $A=0$ for Democrats. In this example, it can be seen that Republicans (and Democrats) are more connected to each other rather than with bloggers with different political opinions. As a direct consequence, one can expect that node embeddings will reflect this separation, making a link prediction or node classification model prone to be biased. For instance, the former model will exclusively connect bloggers from the same political groups, resulting in an emphasis on this cleavage as illustrated in Figure \ref{fig:assortativity_coefficient} (b)). On the other hand, when communities are independent of the protected attribute (Figure \ref{fig:assortativity_coefficient} (c)(d)), 
one can see that the graph does not present such a risk. 

Let us now formally define the assortative mixing coefficient. First, we consider the following quantity:
$$e_{ij} = \frac{card\{(v,u)\in\mathcal{E}; A_{v}=i, A_{u}=j\}}{m},$$ where $m$ is the number of edges in the network. 
Note that when $i\neq j$, $e_{ij}$ corresponds to the fraction of edges in-between nodes having different values for their sensitive attributes, while when $i=j$, it corresponds to the fraction of edges in-between nodes belonging to the same protected group. The matrix defined by $e_{ij}$'s satisfies the property $\sum_{i,j}e_{ij}=1$, and we define $a_i=\sum_{j}e_{ij}$ and $b_{j}=\sum_{i}e_{ij} $ that describes the proportion of edges starting from and ending at each of the attribute values. 
Finally, the assortative mixing coefficient is defined by:
\begin{equation}\label{eq:r}
r = \frac{\sum_{i}e_{ii}-\sum_{i}a_ib_i}{1-\sum_{i}a_ib_i}.
\end{equation}
The coefficient $r$ takes values in $[-1,1]$, where $1$ corresponds to the perfectly assortative case, i.e. nodes with the same value for $A$ exclusively connect with each other; $-1$ corresponds to the disassortative case, i.e. nodes only connect with nodes having a different value for $A$. A \textit{fair} graph will have a mixing coefficient of 0. This coefficient $r$ can be simply computed on any graph to assess if the graph is biased or not \citep{laclau2020optimal}, or used to enforce fairness in the graph \citep{spinelli2021}.

\begin{figure}[!ht]
    \centering
    \subfloat[ ]{ \includegraphics[width=0.3\linewidth]{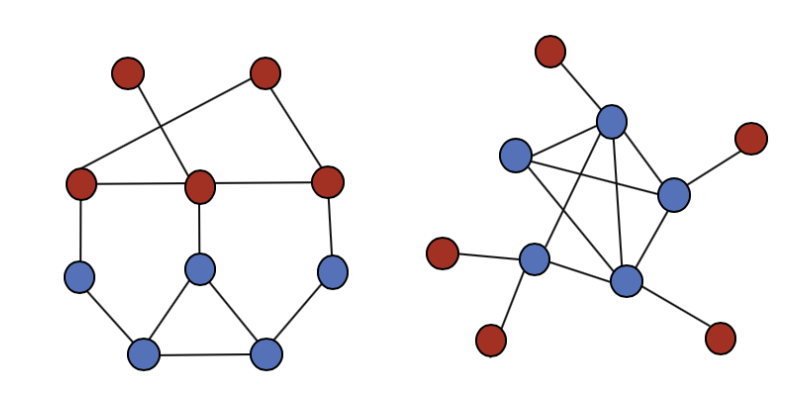}}
    \hspace*{1cm}
     \subfloat[ ]{ \includegraphics[width=0.5\linewidth]{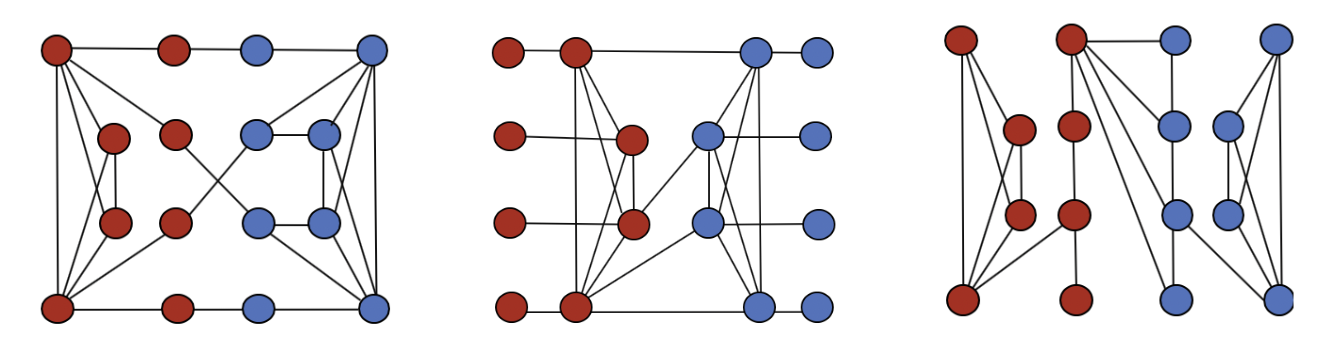}}\caption{\charlotte{Illustration of the concept of Information Unfairness - taken from \cite{jalali2020}. Node color indicates their sensitive group. (a) The left network has an IU of 0.047, while the right network has an IU of 0.1.  It is clear that for the right network, very little information flows between red nodes. (b) Graphs with the same mixing coefficient, but different IU scores.}}
    \label{fig:iu}
\end{figure}

\charlotte{In the same spirit of characterizing the bias of a whole graph, in \citep{jalali2020} the authors proposed the Information Unfairness criteria defined as follows. 
\begin{definition} Given a graph $\mathcal{G}$, $p\in [0, 1]$ a propagation probability for each edge (the same for all edges if no prior is available), $\ell$ the maximum length cascade considered, IU is given by
  $$ IU_{\mathcal{G}, p, l}= max({Dist\{d(f_1g_1), d(f_2g_2): f_1, f_2, g_1, g_2 \in \{1, \cdots, k\}\}}).$$
  where Dist is a distance between two distributions and $d_{f,g}$ is the flow between the pairs of nodes $(u,v)$ such that $A_u = f$ and $A_v = g$.  
\end{definition}
Unlike the mixing coefficient, this metric considers fairness through the lens of information access and captures a different type of bias than the measure presented above. We illustrate the intuition behind this metric in Figure \ref{fig:iu}.
A high IU indicates a lack of information access and hence an exacerbated risk of bias.}

\subsection{Representation Fairness} In this section we describe fairness metric independent from the task at hand and only acting on the learned node or graph embeddings. Representation Bias (RB), originally proposed by \cite{bose2019compositional}, and then coined with the term RB by \cite{buyl2020}, is used for evaluating the bias at the node embedding level. The underlying idea is to consider $A$ as the target variable. Given a node embedding vector as input, RB computes the weighted average over the one-versus-rest AUC scores obtained from the output $\mathbb{P}_h(a, z_v)$ of a classifier $h$ trained to predict the protected attribute $A$ \cris{from the embedding vector $z_v$ of node $v$}. Defining by $V_{a}=\{v|A_v=a\}$ the set of nodes taking the value $a$ for the protected attribute, the RB score is given by:
\begin{equation}
    RB = \sum_{a=0}^k\frac{1}{|V_{a}|}\text{AUC}(\{\mathbb{P}_h(a, z_v)|\forall v\in V_a\}).
\end{equation}
RB $\in [0,1]$ and is ideally close to $0.5$ meaning that the classifier learned from the node embeddings makes random predictions for the sensitive attribute. Compared to the other metrics, RB is not a dyadic-fairness criterion as it focuses on each node individually, and is therefore not sufficient to evaluate the fairness for edge prediction.

\subsection{Fair Prediction on Graphs}
We are now ready to present metrics that are task-dependent. In this part, we focus on node classification and link prediction. 

\charlotte{\paragraph{Node Classification} Fair node classification is the task that shares the most similarity with fair classification on tabular data. The main difference is that we do not directly consider the node in the graph as equivalent to an individual, but rather its learned latent representation.  As a result, we can directly use the metrics proposed in the literature on fair classification on tabular data. The most common are the Disparate Impact (DI), the statistical parity and the Equal Opportunity (EO) \cite{hardt2016equality}.
\begin{definition}
Given a graph $\mathcal{G}=(\mathcal{V}, \mathcal{E})$ and a function $h:\mathbb{V} \rightarrow \{0,\cdots,r\}$, the disparate impact is defined as
$$\text{DI}(h,\mathbb{V},A) = \frac{\mathbb{P}(h(v_i)=\ell|A_{v_i} = 1)}{\mathbb{P}(h(v_i)=\ell|A_{v_i} = 0)},$$
where $A_{v_i} = 1$ indicates that node $v_i$ belongs to some sensitive group. A DI closes to $1$ means that the prediction given by $h$ our classifier is independent from the sensitive attribute.
\end{definition}
}

\charlotte{Statistical parity captures the same information as the disparate impact but the two quantities of interest from the definition of this latter are posed in an equality. However, EO is different from the aforementioned metrics as it acknowledges that in many scenarios, the sensitive characteristic may be correlated with the target variable. Consequently, EO tests whether, given a preferred label $y=1$ (one that confers an advantage) and the protected attribute, a classifier is equally good at predicting that preferred label for all values of that attribute. In other words, EO measures whether nodes who should qualify for an opportunity are equally likely to do so, regardless of their group membership.
\begin{definition}
    In the case of a binary protected attribute, Equal Opportunity (EO) in the context of fair node classification is given by 
    $$P(h(v_i)=1|y_{v_{i}}=1, A_{v_{i}}=1)=P(h(v_i)=1|y_{v_{i}}=1, A_{v_{i}}=0)$$
\end{definition}
We are now ready to present metrics dedicated to evaluating fair edge prediction. 
}

\paragraph{Link prediction} Metrics to measure the fairness of link prediction are often referred to as dyadic fairness metrics in the literature. These metrics consider the bias at the relational level, i.e., the edges. As a consequence, most of the criteria are properties of the joint distribution of the sensitive attribute $A$, the target variable $y$, and the learned edge predictor $h$. 

The first set of metrics are extensions of the aforementioned metrics defined for node classification. In \cite{laclau2020optimal}, the authors considered the following definition for the Disparate Impact for edge prediction. 
  
\begin{definition}
Given a graph $\mathcal{G}=(\mathcal{V}, \mathcal{E})$ and a function $h:\mathbb{V} \times \mathbb{V} \rightarrow \{0,1\}$, the disparate impact is defined as
$$\text{DI}(h,\mathbb{V},A) = \frac{\mathbb{P}(h(v_i,v_j)=1|A_{v_i} \oplus A_{v_j} = 1)}{\mathbb{P}(h(v_i,v_j)=1|A_{v_i} \oplus A_{v_j} = 0)},$$
where $\oplus$ stands for the XOR operator, and $A_{v_i} \oplus A_{v_j} = 1$ (resp. $A_{v_i} \oplus A_{v_j} = 0$) denotes the fact that nodes $v_i$ and $v_j$ belong to the same group (resp. different groups). A DI closes to $1$ indicated a \textit{fair} situation.
\end{definition}
One may note that the difference from the definition considered in node classification  with $\ h:\mathbb{V} \rightarrow\{0,1\}$, comes from the fact that we deal with tuples and implicitly attribute a sensitive variable defined by $A_{v_i} \oplus A_{v_j}$ to each pair of nodes or, equivalently, to an edge.  
Let us consider the job market problem, one can consider a fair node classification task, where the goal is to predict the job category of a given candidate, independently from the gender of the candidate. In that case, the DI corresponds to the ratio of $P(h(v_i)=nurse|A=female)$ and $P(h(v_i)=nurse|A=male)$ for instance, and a fair situation corresponds to the equality of these two quantities, i.e., DI=1. Now, let us consider a slightly different but common task for online professional networks: candidate-candidate matching, i.e., recommending to a given user whom to connect with. In that context, the input consists of two users, hence our objective is to ensure an equal probability of connecting two users independently from the fact that they are both men or  women (denoted by $A_{v_i} \oplus A_{v_j} = 0$). In both cases, the value of $DI$ has the same interpretation. 

In the same spirit, one can extend the notion of statistical parity and equal opportunity for edge prediction, as follows. 
\begin{definition} Given a graph $\mathcal{G}=(\mathcal{V}, \mathcal{E})$ and a function $h:\mathbb{V} \times \mathbb{V} \rightarrow \{0,1\}$, statistical parity for an edge predictor $h$ on $A$ with respect to $\mathbb{V}$ is given by: 
    $$\mathbb{P}(h(v_i,v_j)=1|A_{v_i} \neq A_{v_j}) = \mathbb{P}(h(v_i,v_j)=1|A_{v_i} = A_{v_j})$$
    or equivalently
    $$\mathbb{P}(h(v_i,v_j)=1|A_{v_i} \oplus A_{v_j} = 1) = \mathbb{P}(h(v_i,v_j)=1|A_{v_i} \oplus A_{v_j} = 0),$$
	\label{def:stat_parity}
	where $\oplus$ stands for XOR operation as stated in the definition of DI. 
\end{definition}

\begin{definition}Equal Opportunity (EO), in the context of edge prediction can be formulated as follows.
    \begin{equation*}
    P(h(v_i,v_j)=1|y(v_i,v_j)=1, A_{v_i}=A_{v_j})
= P(h(v_i,v_j)=1|y(v_i,v_j)=1, A_{v_i}\neq A_{v_j}).
\end{equation*}
In the literature, this measure is usually expressed as a difference:
\begin{equation}\label{eq:eo}
    \Delta EO = |P(h(v_i,v_j)=1 | y(v_i,v_j)=1,A_{v_i}=A_{v_j}) - P(h(v_i,v_j)=1 | y(v_i,v_j)=1,A_{v_i}\neq A_{v_j})|.
\end{equation}
Note that $\Delta EO \in [0, 1]$, where $0$ represents a fair situation.
\end{definition}
Going back to our candidate-candidate matching task, statistical parity focuses on having the positive outcome independent of the protected class, e.g, $P(h(v_i, v_j)=follow |(A_{v_i}=female, A_{v_j}=male))=P(h(v_i, v_j)=follow |(A_{v_i}=male, A_{v_j}=male))$. Equal Opportunity requires the outcome to be independent of the protected class $A$, conditionally on the actual label $y$. A direct consequence of this additional assumption is to reduce the risk of blindly predicting links between users to fill in the gap between two protected groups. 

Finally, in \cite{rahman2019fairwalk}, authors proposed to define the Acceptance Rate Parity (ARP). This metric is a ranking-based measure introduced to evaluate fair edge prediction. 

\begin{definition}
The acceptance rate accounts for the relative frequency for which a link between two nodes with different 
attributes $A_{v_i}=a$ and $A_{v_j}=b$ appears in the overall top-k highest prediction scores. For a given node $v_i$ the list of its link recommendations is denoted by $\rho_{v_i}$, and $v_j \in \rho_{v_i}$ indicated that a link has been predicted between $v_i$ and $v_j$.
\begin{equation}
    \alpha_{x,y}=\frac{1}{N_{x,y}}|\{(v_i,v_j)|v_j \in \rho(v_i), v_i \in V_x, v_j \in V_y\}|
\end{equation}
where $N_{a,b}$ represents all possible combinations between the nodes having attribute values $a$ and $b$ and can be given as,
\begin{equation*}
    N_{a,b} = |\{(v_i,v_j)| v_i \in V_a, v_j \in V_b\}|.
\end{equation*}
A high ARP indicates a fair situation. Note that this metric depends on the length of the considered list of predictions for a given node. 
\end{definition}
Let us illustrate this metric with our previous example. We first consider, a given user $v_i$, with protected attribute $A_{v_i} = female$, then look at the attributes of the top-5 candidates recommended to this particular user: $(A_1, A_2, A_3, A_5)=female$, $A_4 = male$. The ARP counts the relative frequency of males in the top-5, here $\frac{1}{5}$, meaning that the model decision is unfair. While ARP is intuitive and easy to interpret, it presents the same weakness as the DI or statistical parity, as it does not account for the relevance of the prediction along with the protected attribute. Hence, models that are simply moving up male in the top-5 list regardless of their relevance for that particular user will be considered fair, while they negatively impact individual fairness notions.

\subsection{Counterfactual Fairness}
We are now ready to present metrics proposed to account for counterfactual fairness. To the best of our knowledge,  \cite{agarwal2021towards} were the first ones to address this challenge. To proceed, they define two metrics to assess the level of counterfactual (un)fairness which they relate to a stability issue. The first one referred to as the unfairness score corresponds to the percentage of test nodes for
which predicted label changes when the node’s sensitive attribute is flipped. The second one, referred to as the instability score represents the
percentage of test nodes for which predicted label changes when random noise is added to node-sensitive attributes.

\subsection{Measuring Individual Fairness}
All aforementioned metrics are defined with respect to some sensitive attribute(s). Different from these latter, intuitively individual fairness aims to ensure that two individuals (in this particular case, two nodes) sharing similarities should receive similar treatments. In our context, the term \textit{treatment} might refer to different processes. For instance, for node classification, \textit{treatment} refers to the labeling process and individual fairness means that we expect that two similar nodes should be predicted with the same label. And as we go backward in the general process described in Figure \ref{fig:process}, the same expectation goes for node embeddings or any function taking the graph as input.
This intuition was formally captured in the seminal work of \cite{DworkHPRZ12} and extended to graphs by \cite{laclau2020optimal}. 
\begin{definition} A mapping $\phi:\mathbb{V} \rightarrow \mathbb{V}$ satisfies the $(D,d)$-Lipschitz property if for every $v_i,v_j \in \mathbb{V}$, and two metrics $D, d:\mathbb{V}\times \mathbb{V} \rightarrow \mathbb{R}^+$, we have:
\begin{align*}
    D(\phi(v_i), \phi(v_j)) \leq d(v_i,v_j),
\end{align*}
\end{definition}
The intuition behind such a definition is to say that any transformation or mapping of the original graph (here described by all pairs of nodes) should respect the similarity between them after the mapping.  It follows naturally from this formalization, that the major concern of individual fairness lies in the requirement of an appropriate similarity measure which often requires the resolution of a non-trivial problem that may need expert knowledge and certainly depends on the application domain initially targeted. Building upon this definition, several metrics have been proposed to measure the level of individual fairness of machine learning models on graphs. In \cite{kang2020inform, dong2021}, the authors chose $D(.,.)$ to be the Euclidean norm and compare the use of either the Jaccard index or the cosine similarity for $d(.,.)$. It is worth noting that while they used this principle to enforce individual fairness, they did not consider this definition to evaluate any individual fairness performance, relying instead on a comparison between the utility (AUC) of an original model and the utility of a fair version of that model. In contrast, \cite{laclau2020optimal} adopts the definition of consistency originally proposed by \cite{zemel2013learning} to assess individual fairness in the context of representation learning as follows: 
\begin{definition}
    Consistency consists in comparing the model output for a given node $v_i$ to its k-Nearest Neighbors $kNN(v_i)$. For graphs,  $kNN(v_i)$ is approximated with $kNN(z_{v_i})$, the k-Nearest Neighbors in the embedding space. 
    $$Consistency = 1-\frac{1}{nk}\sum_{i=1}^{n} |h(z_{v_i})-\sum_{j\in kNN(z_{v_i})}h(z_{v_j})|,$$
where $n$ is the number of nodes in the graph, $k$ is the number of neighbors considered for each node and $h$ is a prediction function (e.g. node classifier). A perfectly fair model is one with a consistency of $1$. 
\end{definition}

The following sections present the main algorithmic contributions for fair graph learning. Section \ref{sec:preproc} covers pre-processing approaches that aim to transform the original data, Section \ref{sec:fair_embedding}, in-processing methods by distinguishing that build upon existing node embedding-based models from methods that aim to solve the problem in an end-to-end fashion. Finally, even if this approach is much less common, the last strategy consists to remove the bias during the post-processing. Section \ref{post-processing} gives  examples of this type of method.

\section{Repairing Bias at the Origin}\label{sec:preproc}
\cris{In our typology, the first family of methods aims at removing bias from the graph structure itself during the pre-processing step, by focusing on modifying directly the graph (i.e. the adjacency matrix)}. The major advantage of these models is that they are embedding-agnostic. Among these methods, we present in more detail
\texttt{MaxFair} \cite{jalali2020},  \texttt{FairOT} \citep{laclau2020optimal} and  \texttt{FairDrop} \citep{spinelli2021}. 

\cris{Modifying the graph structure can be achieved by edge rewiring, by either adding, deleting or redistributing edges.  In \texttt{MaxFair} \cite{jalali2020}, the authors characterize the flow of information between groups of users, defined according to the sensitive attribute, using joint accessibility distribution. They introduce the IU metric defined in Section \ref{sec:metrics} and their strategy consists in adding edges in the network so as to reduce this score. Intuitively the main idea is to produce a graph with a more equal flow of information between and within the groups.}

\smallskip

In  \texttt{FairOT} \citep{laclau2020optimal}, the authors propose to \textit{repair} the original graph using optimal transport.
To this end, they propose to cast the problem of debiasing the graph as a problem of alignment between node distributions of nodes belonging to different protected groups, where the latter are taken to be the rows of the normalized adjacency matrix. Their proposition is derived from a theoretical analysis of the disparate impact. In addition, their approach is the first one that allows explicit control of the
trade-off between individual and group fairness through a Laplacian regularization term added to the optimal transport objective. The model is evaluated on both synthetic graphs and real-world benchmarks for the task of edge prediction.

\texttt{FairDrop} \citep{spinelli2021} is also a pre-processing technique that modifies the adjacency matrix during a training step to compensate for the homophily due to the sensitive attribute. This approach is based on a biased edge dropout which consists at each step of the training to remove edges between the linked nodes according to a randomized response mechanism that cuts more links between nodes having the same sensitive attribute value. The authors evaluate \texttt{FairDrop} on link prediction, both through a random walk model for building node embeddings and in a graph convolutional network able to solve the task in an end-to-end fashion. Compared with some in-processing methods, the experimental results obtained on traditional benchmark show that \texttt{FairDrop} decreases the RB with a negligible drop in accuracy. 

\cris{A clear advantage of these methods which repair the graph during a pre-processing step is that they can be combined with any embedding models or graph-based machine learning techniques, which makes them quite flexible. However, one should bear in mind that in this case, nothing prevents node embedding models or classifiers used afterward from retrieving signals about the protected attribute.}
\smallskip

\cris{The second type of approach is precisely based on embedding models. It encodes parts of the graph into a low-dimensional real space while imposing fairness constraints. Then, these vector representations can be exploited by classical machine learning methods designed for tabular data, for solving the downstream tasks, for instance, node classification or link prediction. Graph neural network networks are also popular embedding models that simultaneously allow handling the task. Next, we present all these models in the family of in-processing methods.}

\section{Learning fair node embedding}\label{sec:fair_embedding}
\charlotte{
In this section, we differentiate contributions to the domain of fairness for graphs between, unsupervised node embedding approaches, that build upon existing node embedding models, from models which are solving the problem in an end-to-end fashion, most of them being based on graph neural networks.} 
\subsection{Fairness based on Unsupervised Node Embedding}
To the best of our knowledge, the first model proposed to address the problem of fairness for graphs is \texttt{Fairwalk}~\cite{rahman2019fairwalk}: a fair extension of \texttt{Node2vec}. \texttt{Fairwalk}, relies on a modification of the random walks to induce more fairness. This method modifies the transition probability of \texttt{Node2vec} for the generation of unbiased traces and can be detailed in the following two steps.
    
\begin{enumerate}
    \item First, a corpus of traces is generated by performing random walks. Formally, denoting by $c_i$ the $i$-th node in a given walk, the next node is selected among all neighbors of $c_i$, i.e., 
    $$\mathbb{P}(c_{i+1} = v|c_{i} = u) = \left\{
    \begin{array}{ll}
        \frac{1/a}{|A_{N_{u}}^{a}|} & \mbox{if } A_{v}^{a} = 1 \mbox{ and } \{u,v\} \in \mathcal{E} \\
        0 & \mbox{otherwise,}
    \end{array}
    \right.
$$
where $a \in \{1,\cdots, k\}$ denotes the modality of the sensitive attribute $A$, $A_{N_{u}}^a$ is the number of nodes in the neighborhood of $u$ belonging to the group $a$ and $A_{v}^{a} = 1$ indicates that node $v$ belongs to the $a$-th group of the sensitive attribute. As a result, each generated random walk has a higher probability to contain nodes of different groups.
    \item Then, \texttt{Fairwalk} uses the generated corpus to learn the embedding vectors through a SkipGram architecture that maximizes the log-probability of observing a network neighborhood for a node conditioned on its feature representation: 
    $$\argmax{Z}\prod_{u \in \mathcal{V}}\prod_{v \in N_u} \mathbb{P}(v|z_u).$$
\end{enumerate} 

\texttt{Fairwalk} works well on graphs in which the neighboring nodes have different sensitive attributes, for example, the graph $G_2$ presented in Figure \ref{fig:assortativity_coefficient}(c). However, one can expect this model to fail in extreme cases such as the graph $G_1$ presented in Figure \ref{fig:assortativity_coefficient}(a), as during the random walks, only the nodes belonging to the same group will be discovered in the direct neighborhood of a given node.

In the same spirit, \texttt{CrossWalk} \cite{khajehnejad2021crosswalk} is based on a re-weighting procedure for building the random walks and can therefore be used with any random walk-based algorithms including \texttt{DeepWalk} \cite{perozzi2014deepwalk} and \texttt{Node2vec} \cite{grover2016node2vec}, to name a few. 
\texttt{CrossWalk} aims to address the shortcomings of \texttt{Fairwalk} by assigning more weights to both edges that connect nodes on the boundary of the protected groups, and edges connecting nodes from different groups. It has been evaluated in various graph-related tasks such as link prediction, node classification and influence maximization. and iy obtained superior results than \texttt{Fairwalk}, resulting from the group peripheries exploration.

\smallskip

Different from the aforementioned techniques, \texttt{Debayes} \cite{buyl2020} is a Bayesian approach based on the Conditional Node Embeddings (\texttt{CNE}) \cite{kang2019conditional} method, where the sensitive information is modeled in the prior distribution. Given a graph $\mathcal{G}=(\mathcal{V}, \mathcal{E})$, \texttt{CNE} finds an embedding $\textbf{Z}$ by maximizing $P(\mathcal{G}|\textbf{Z}) =  \frac{P(\textbf{Z}|\mathcal{G})P(\mathcal{G})}{P(\textbf{Z})}$. The prior knowledge about the node degree is modeled by the prior distribution $P(G)$, expressed by the following constraint:
\begin{equation}
\sum_{v \in V}P((v,v') \in E) = \sum_{v \in V} \mathds{1} \left((v,v') \in \mathcal{E}\right) .
\label{CNEeq1}
\end{equation}
\texttt{Debayes} extends \texttt{CNE}, with a prior to model the sensitive attribute by replacing the constraint \eqref{CNEeq1} with:
\begin{equation*}
\sum_{v \in V_s}P((v,v') \in E~|~A_v = a)= \sum_{v \in V_a} \mathds{1} \Big((v,v') \in \mathcal{E}\Big),
\end{equation*}
where $V_a = \{v | A_v = a\}$. With this prior, debiased embeddings, containing minimal information about sensitive attributes, are obtained during the training step. 

\subsection{End-to-end Fairness}

\charlotte{In this section, we present approaches designed to solve the problem of fair learning on graphs in an end-to-end style, meaning that node embeddings and the model to solve the task at hand, are learned simultaneously. In this context, Graph Neural Networks (GNNs) models have recently established themselves as state-of-the-art for many graph-related downstream tasks, including node classification and edge prediction. These models usually consider two types of input: the graph structure (usually in the form of an edge list or the adjacency matrix), and a matrix of node and/or edge attributes. GNNs are able to leverage both types of information to learn powerful representations, and by design, model the homophily principle: for given node attributes, similar nodes are more likely to be connected to each other than dissimilar ones. While this assumption is valid in many real-world scenarios, homophily has been identified as a potential source of bias in social networks. As a result, in recent years much of the effort has been focused on debiasing GNNs. We now present these approaches, and we classified these latter depending on the task they both propose to solve and have been evaluated for. Once again, we focus on node classification and edge prediction, but the reader should be aware that other tasks have been considered in the literature, including node ranking \cite{dong2021}, influence maximization \cite{TsangWRTZ19, RahmattalabiVFR19, RahmattalabiJLV21, khajehnejad2021, AliBCMGS23}, and community detection \cite{KleindessnerSAM19}.}

\subsubsection{Node Classification}

\begin{figure}[!t]
    \centering
    \subfloat[\texttt{MONET}]{\includegraphics[width=0.4\linewidth]{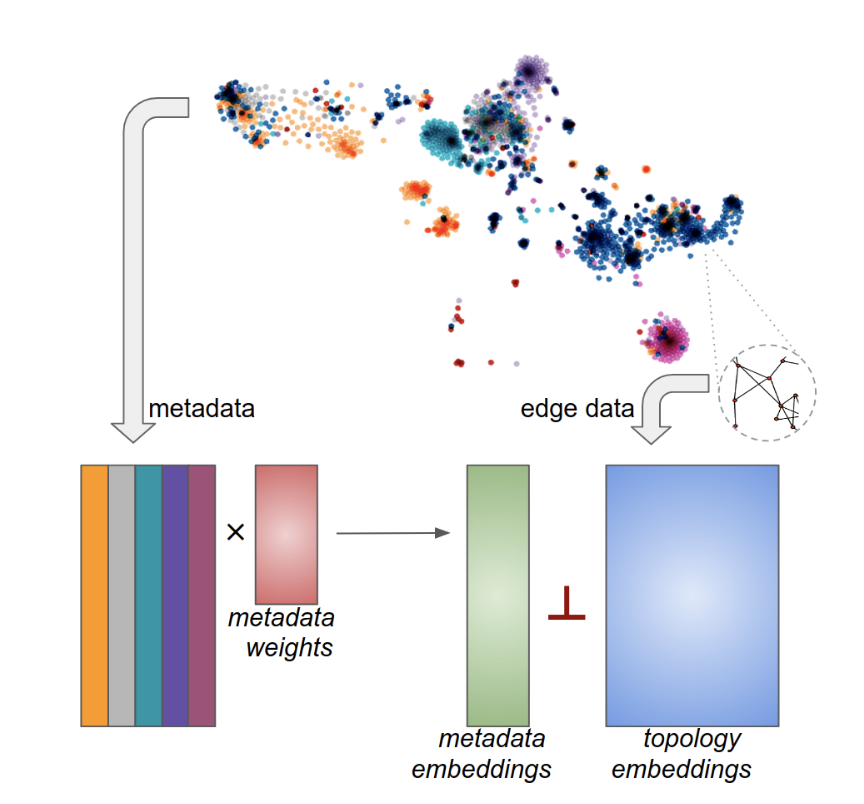}}
    \hspace*{1.3cm}
    \subfloat[\texttt{NIFTY}]{\includegraphics[width=0.4\linewidth]{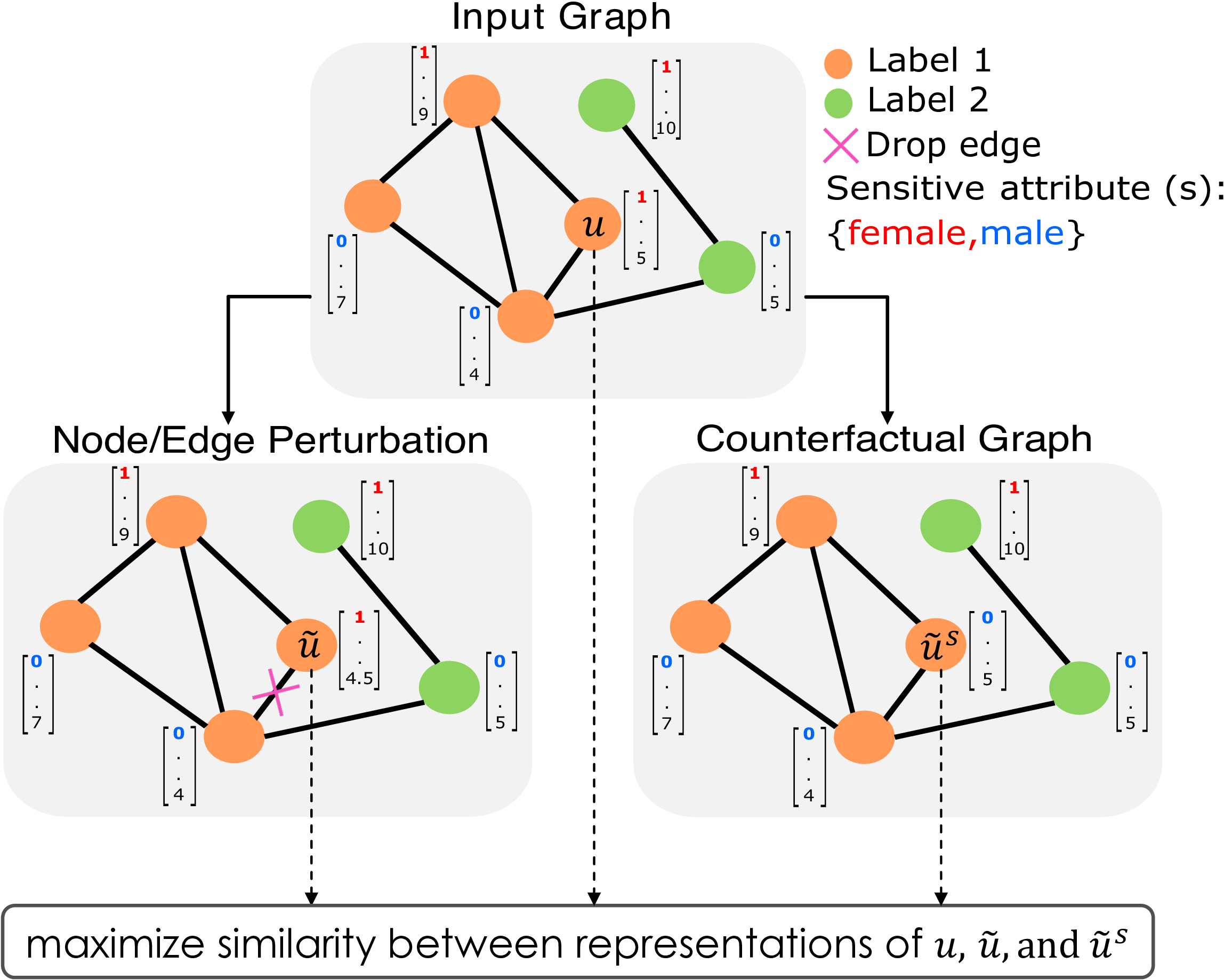}}
    \caption{Illustration of two GNN-based models for graph fairness: (a) \texttt{MONET}\citep{palowitch2020monet} which imposes independence (through orthogonality constraints) between the embedding spaces of features and topology; 
    and (b) \texttt{NIFTY} \citep{agarwal2021towards} that generates perturbated and counterfactual graphs to ensure stable node representation. Images are taken from original papers.}
    \label{fig:fair-gnn}
\end{figure}

The first proposition in that direction is the Metadata-Orthogonal Node Embedding Training (\texttt{MONET}) model  \cite{palowitch2020monet}, which performs debiasing of node embeddings at training time with a neural network component coined with the term \texttt{MONET} unit. The model learns jointly a graph topology embedding matrix 
and a graph metadata embedding matrix while enforcing linear independence between the two embedding spaces through Singular Value Decomposition
(SVD). The prime limitation of this approach relies on the fact that it only performs a linear debiasing and as a result, the embeddings learned with \texttt{MONET} are only fair with respect to linear models for the edge prediction task. Thus, non-linear models (e.g., support vector machines with non-linear kernels) are able to retrieve part of the bias directly from the embeddings. The performances of the model are evaluated, in terms of representation bias, by considering node classification on the sensitive attributes and on  shilling attacks where several users act together to artificially increase the likelihood that a particular influenced item will be recommended for a particular target item. An illustration of the proposal is given in Figure \ref{fig:fair-gnn}(a).

\smallskip

In \citep{agarwal2021towards}, the authors focus on two weaknesses of GNNs: stability and fairness, where the latter is defined as counterfactual fairness. Counterfactual fairness accounts for robustness w.r.t modifying the sensitive attribute while stability accounts for robustness w.r.t perturbing the node attributes and/or link information. To proceed, they propose a new versatile framework, named \texttt{NIFTY} (uNIfying Fairness and stabiliTY), that has the ability to encompass several GNN architectures. \texttt{NIFTY} proposes to modify both the objective loss and the architecture of the GNNs to tackle the problems of stability and fairness. It relies notably on the generation of augmented views of the original graph with counterfactual nodes and perturbated edges and the Lipschitz constant to develop layer-wise weight normalization. Overall, it demonstrates superior performance by a large margin than its competitors. They studied two versions of \texttt{NIFTY}: the first on is based on Graph Convolutional Network \citep{kipf2016} and the second one is based on GraphSage \citep{hamilton2017b}. An illustration of the general idea of \texttt{NIFTY} is provided in Figure \ref{fig:fair-gnn}(b).

\smallskip

\texttt{FairGNN} \cite{enyan2021} focuses on the problem of learning fair Graph Neural Network (\texttt{GNN}) representations in the presence of partial sensitive attribute information. The authors consider the problem of node classification and propose to mitigate bias by using adversarial debiasing on the final layer of the \texttt{GNN} i.e., the learned node representations. The main drawback of this model is its limitation to the node classification task. \cris{In the same spirit, 
noting that attributes highly correlated to the sensitive attribute can propagate bias to learned representations and decisions made by a classifier, \texttt{FairVGNN}\cite{wang2022b} incorporates an adversarial debiasing for automatically masking both sensitive and correlated features and alleviate discrimination. Evaluated with different GNN-backbones (GCN, GIN, and GraphSAGE) and compared against NIFTY \citep{agarwal2021towards}, EDITS \cite{DongLJL22} and FairGNN \cite{enyan2021}, \texttt{FairVGNN} achieves a correct trade-off between fairness and performances for node classification.
}

An important effort has also consisted in relaxing a strict definition of fairness by adding one or multiple regularisation terms in the loss function optimized by the GNN. \cris{For instance, \texttt{DFGNN} \cite{OnetoND20} integrates fairness in GraphSAGE through the use of two regularizers imposing demographic parity constraints based on a convex relaxation or a reformulation of this fairness constraint in terms of Sinkhorn distance.}
Similarly, assuming that the labels of adjacent nodes are likely to be the same, \texttt{FGNN} \cite{zhang2022} searches a trade-off between accuracy and fairness but in the context of semi-supervised learning where labeled and unlabeled nodes are provided. The authors propose two models: the first one, FSMC  is a semi-supervised margin classifier that relies on the optimization of an objective function including a loss for both the classifier and label propagation, and fairness regularizers over labeled and unlabeled data. The second,  FGNN, built on GNNs, with classification loss and fairness loss,  learns fair representation for both types of nodes and assign a  label to the unlabeled nodes.
\smallskip

Finally, we will end this section dedicated to node classification by citing FairHIN proposed by Zeng et al. who use orthogonal projection  to decorrelate the node embeddings from the sensitive attribute  in heterogeneous networks \cite{zeng2021}. 

\smallskip
\subsubsection{Edge Prediction}
\texttt{FairEGM} \cite{CurrentHG022}  mitigates the bias learned by the GCN by optimizing a loss function integrating a demographic parity constraint. This approach emulates adding nodes and reweighting edges but without the computational cost of such modification of the graph structure. FairEGM performs better than FairWalk and FairAdj on the classical benchmarks CiteSeer, Cora, Facebook and, better than FairAdj on Pubmed in terms of demographic parity but, this is at the cost of a loss of performance in terms of prediction compared to a simple graph convolution network.
The work of \cite{li2021} relies on the same hypothesis that bias is inherent to the graph topology and exploited by pre-processing models and \texttt{Crosswalk}. The authors propose to modify the adjacency matrix of the graph while training node embeddings with a GNN by modifying the message-passing strategy. They notably demonstrate that fair vertex representations are a sufficient condition to achieve demographic parity (group fairness) in link prediction. To proceed they propose a reweighing schema of the edges (intra and inter) that guarantees a reduction of the representation discrepancy between two sensitive groups. They implement their proposal within the framework of variational graph auto-encoders \citep{kipf2016} and show that it can outperform adversarial training with various fairness-accuracy trade-offs. 

\smallskip

\cris{Another recent approach that falls into this category is the Unbiased Graph Embedding \texttt{UGE} \citep{wang2022} framework that relies on similar principles that the aforementioned ones. Indeed \texttt{UGE} generates a new bias-free graph by reducing the dependence between the structure of the observed graph and the effect of the sensitive attribute. Like in \texttt{FairEGM} \cite{CurrentHG022}, previously cited, unbiased graph embeddings  are obtained by adding a regularization term to the loss function to enforce  group fairness. The ability of \texttt{UGE} to produce unbiased embeddings as well as the utility of these embeddings for link prediction are then evaluated using well-known graph neural network architectures.}

\smallskip

\cris{\texttt{HM-EIICT} \cite{SaxenaFP22} applies in a different context, that of the prediction of intra- and inter-community links in a network with a community structure. Starting from the observation that the similarity between a pair of nodes belonging to the same community is greater than that of nodes belonging to different communities, \texttt{HM-EIICT} estimates threshold values on a learning sample or after having identified the community structure and, then it integrates these thresholds into the calculation of the similarity measure used to decide whether or not two nodes belong to the same community. Although this method has been presented as a way to reduce bias, it should be noted that only its accuracy in terms of prediction has been evaluated.}
Finally Fairness Regularizer (\texttt{FIPR}) \cite{buyl2021} generalizes the idea of \texttt{DeBayes} by proposing a regularization term to encourage fair link prediction that can be applied with any probabilistic network models. This regularizer is defined by the KL-divergence between the probabilistic node embedding model and its I-projection. One of the strengths of \texttt{FIPR} is its ability to be flexible both with respect to the node embedding approach but also with respect to the fairness metric.
    \begin{figure*}[!ht]
    \centering
    \begin{tikzpicture}
    \node(input)[minimum width=1.5cm,minimum height=1.5cm,label=below:$G$]at (-1.2
    ,0){\includegraphics[width=0.15\textwidth]{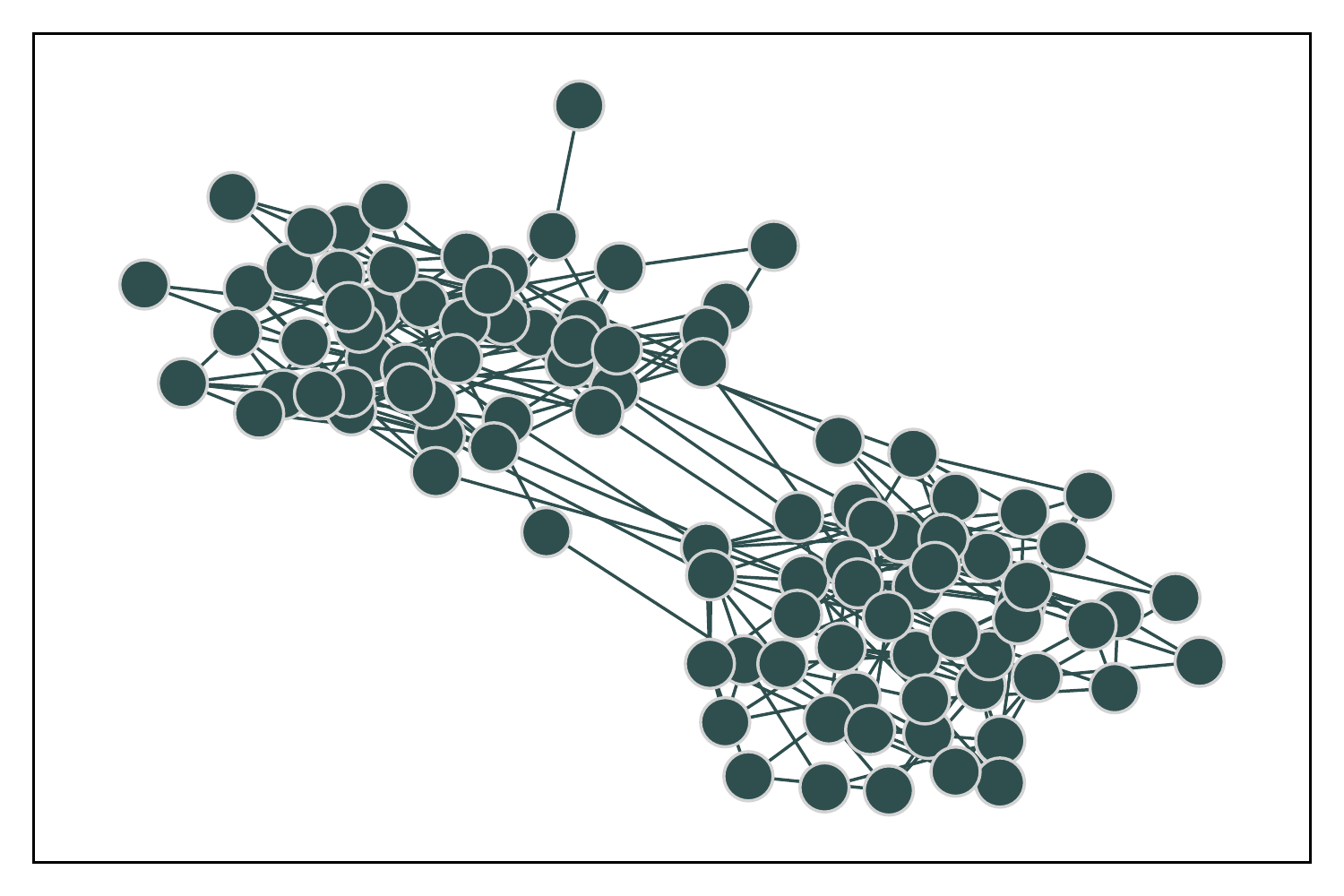}};
    \node [draw=yellow!70, fill=yellow!10,
    minimum width=2cm,
    minimum height=1cm,
    right=.5cm of input
    ]  (generator) {Generator};

    \node (embeddings) [right of=graph,minimum width=.5cm,minimum height=3cm,draw=red!30,label=below:$z_i \in \mathbb{R}^d$, fill=orange!10,right=.01mm of generator] {};
    
    \node [draw= cyan!55, fill = cyan!10,
    minimum width=2.2cm,
    minimum height=.8cm,yshift=-.6cm,
    right=.7cm of embeddings,label=below:$D$
    ]  (discriminator) {Discriminator};
    
    \node [draw= gray!35, fill=gray!10,
    minimum width=2cm,
    minimum height=.8cm,yshift=.6cm,
    right=.7cm of embeddings
    ]  (link_pred) {Link Prediction};
    
    \draw[-stealth] (input.east) -- (generator.west)
    node[midway,above]{};
    
    \draw[-stealth] (generator.east) -- (embeddings.west)
    node[midway,above]{};
    
    \draw[-stealth] (embeddings.east) -- (link_pred.west)
    node[midway,above]{};
    
    \draw[-stealth] (embeddings.east) -- (discriminator.west)
    node[midway,above]{};

    \end{tikzpicture}
    \caption{FLIP Architecture: The generator takes an attributed network as input and learns a representation for each node. FLIP uses DeepWalk \cite{perozzi2014deepwalk} as the generator, other representation learning frameworks can also be used. The Discriminator takes as input the representation for each pair of nodes and tries to predict if it belongs to intra-group or inter-group. The link prediction component takes as input the learned node representations and tries to predict if there is a link.}
\label{fig:flip}
\end{figure*}
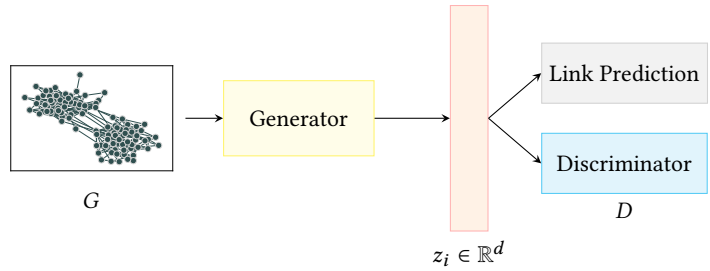

\section{Post-processing Approaches}
\label{post-processing}
\begin{figure}[!ht]
    \centering
    \begin{tikzpicture}
        \node(graph)[minimum width=1.5cm,minimum height=1.5cm,label=below:$G$]at (-6,0){\includegraphics[width=0.15\textwidth]{ill_graph.pdf}};
        
        \node (embedding) [right of=graph,minimum width=.5cm,minimum height=3cm,xshift=.6cm,draw=red!30,label=below:$z_i \in \mathbb{R}^d$, fill=orange!10] at (-5,0) {};
        
         \node (plus) [right of =embedding] at (-3.8, 0) {\large \textbf{+}};
         
        \node (attr) [right of=plus, minimum width=.5cm,minimum height=2cm,draw=gray!30,label=below:$A$] at (-3,0){\includegraphics[scale=.15]{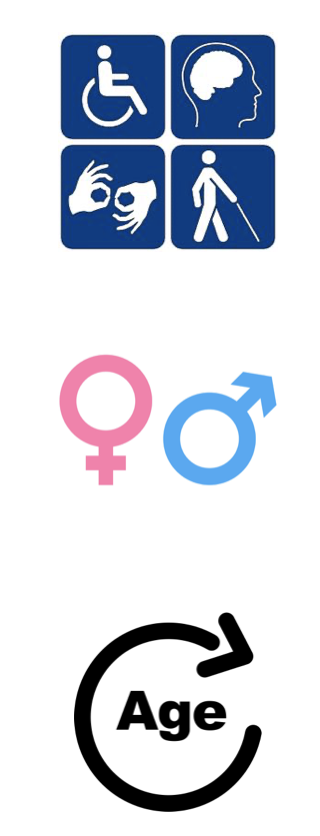}};

        \node (sens_attr) [right of = attr, minimum width=1.6cm,minimum height=3cm,xshift=1.5cm,draw=blue!5, fill=blue!5,label=above:Filters]{};
        
        \node (gender) [right of = attr, minimum width=1.1cm,minimum height=.5cm,xshift=1.5cm,yshift= 0cm, draw=white, fill={rgb:red,1;green,2;blue,5} ]{\textcolor{white}{Gender}};
        
        \node (age) [right of = attr, minimum width=1.1cm,minimum height=.5cm,xshift=1.5cm, yshift=1cm, draw=white, fill={rgb:red,1;green,2;blue,5}]{\textcolor{white}{Age}};
        
        \node (other) [right of = attr, minimum width=1.1cm,minimum height=.5cm,xshift=1.5cm, yshift=-1cm, draw=white, fill={rgb:red,1;green,2;blue,5}]{\textcolor{white}{...}};
        
        \node (filtered) [right of=age,minimum width=.5cm,minimum height=3cm,xshift=.8cm,yshift=-1cm,draw=red!30,label=below:$\tilde{z}_{i}\in \mathbb{R}^d$, fill=orange!10] {};
        
        \node (Discriminators) [right of = filtered, minimum width=1.6cm,minimum height=3cm,xshift=.6cm,draw=black!5, fill=black!5,label=above:Discriminators]{};
        
        \node (d_gender) [right of = filtered, minimum width=1.1cm,minimum height=.5cm,xshift=.6cm,yshift= 1cm, draw=white, fill=black!75]{\textcolor{white}{$D_{G}$}};
        
        \node (d_age) [right of = filtered, minimum width=1.1cm,minimum height=.5cm,xshift=.6cm, yshift=0cm, draw=white, fill=black!75]{\textcolor{white}{$D_{A}$}};
        
        \node (d_other) [right of = filtered, minimum width=1.1cm,minimum height=.5cm,xshift=.6cm, yshift=-1cm, draw=white, fill=black!75]{\textcolor{white}{$...$}};
        
        \draw[-stealth] (graph.east) -- (embedding.west)
        node[midway,above]{};
        \draw[-stealth] (attr.east) -- (age.west)
        node[midway,above]{};
        \draw[-stealth] (attr.east) -- (gender.west)
        node[midway,above]{};
        \draw[-stealth] (attr.east) -- (other.west)
        node[midway,above]{};
        \draw[-stealth] (sens_attr) -- (filtered) node[above,midway]{};
        \draw[-stealth] (filtered.east) -- (d_age.west)
        node[midway,above]{};
        \draw[-stealth] (filtered.east) -- (d_gender.west)
        node[midway,above]{};
          \draw[-stealth] (filtered.east) -- (d_other.west)
        node[midway,above]{};
    \end{tikzpicture}
    \caption{\texttt{CFC}: the graph $G$ comprises of nodes with sensitive attributes such as age, gender, etc. The node embeddings ${z}_{i}\in \mathbb{R}^d$ for $G$ are informative of these sensitive attributes. Compositional Fairness Embeddings $\tilde{z}_{i}\in \mathbb{R}^d$ are generated after filtering out the sensitive information with the help of Discriminators.}
    \label{fig:bose}
    \end{figure}

One of the first approaches in this line of research is 
Compositional Fairness Constraints (\texttt{CFC}) \cite{bose2019compositional} that aims to generate node embeddings that are invariant of the sensitive attributes. To proceed, the authors adapted a two-step procedure consisting  (1) to learn embeddings for each node and (2) to filter out bias from these embeddings. More precisely, they propose to train a set of \textit{filters} (for the sensitive attributes) such that the adversarial discriminators are unable to classify the sensitive attribute from the filtered embedding. Hence, the goal is to minimize the mutual information between the filtered embedding and the protected attributes. These filters are defined for potentially multiple sensitive attributes $m$ as $f_m: \mathcal{R}^{d} \rightarrow \mathcal{R}^{d}$ and are trained to remove the sensitive information about the $m^{th}$ attribute (see an illustration in Figure \ref{fig:bose}). Note that this approach is the only one that can consider several protected attributes simultaneously. Then, the filtered embedding becomes the input to the adversarial discriminators which try to predict the sensitive attribute of the nodes. As a result, this approach particularly focuses on reducing the representation bias (RB) presented in Section \ref{sec:metrics}. However, by processing nodes individually, \texttt{CFC} is not considering the problem of dyadic fairness. 

While \texttt{CFC} considers the representation bias, Fairness-Aware Link Prediction (\texttt{FLIP}) \cite{masrour2020} proposes an adversarial architecture focusing on controlling the impact of graph level metrics for fairness. More precisely, the authors address fairness through the lens of modularity, which in their case is similar to homophily. To proceed, they post-process the prediction output of their model so as to reduce the modularity of the predicted network:  this model encourages predictions of more inter-group links than in the original network.

Finally, we conclude this section with InFoRM \cite{kang2020} even if it differs from the  aforementioned models as the authors propose several metrics for both individual as well as group fairness  and derive three algorithmic contributions to
optimize these metrics: 1) a pre-processing step to debias the input graph; 2) an in-processing step to debias the mining
model; and 3) a post-processing step to debias the model outcome. They instantiated their framework with PageRank \cite{page1999} and LINE \cite{tang2015} for link prediction, and with a spectral clustering model for community detection.

\section{Benchmark Graphs}\label{sec:data}
\url{https://github.com/laclauc/fairness-graph/}.

\subsection{Synthetic Graphs}

Synthetic graphs can allow us to better understand the behavior of a model in a controlled experimental environment and to identify possible weaknesses. We describe hereafter the synthetic scenarios and the generative process associated with them, introduced in \cite{laclau2020optimal}. 

The authors propose to generate five synthetic graphs (\Gu--\Gc) composed of 150 nodes each (note that this number can be easily increased) based on the stochastic block model with different block parameters. This latter is done using the \texttt{networkx} library \footnote{\url{https://networkx.github.io/}}. Following a similar process, we propose to add a sixth scenario, referred to as (\Gs)  in the following. We believe that this scenario could be particularly helpful to the research community, as very few models can handle it. For the sake of reproducibility, we provide some of the parameters used to generate the graphs in Table \ref{tab:parameters}. In addition, we report the average values of the assortativity mixing coefficient obtained over 10 trials. Note that this criterion, as explained in Section \ref{sec:metrics}, amounts to measuring the potential bias that one can expect when dealing with such structure. Each node is associated with one or multiple sensitive attributes. For the first four graphs, this sensitive attribute $A$ is binary, for {\Gc}  $A$ is a multiclass attribute taking three modalities \cris{($k=3$)}. Using the same protocol, we propose {\Gs}, which corresponds to a graph with two binary protected attributes $A$ and $A'$.

\begin{figure}[!t]
    \centering
    \subfloat[\Gu]{\includegraphics[width=0.31\linewidth]{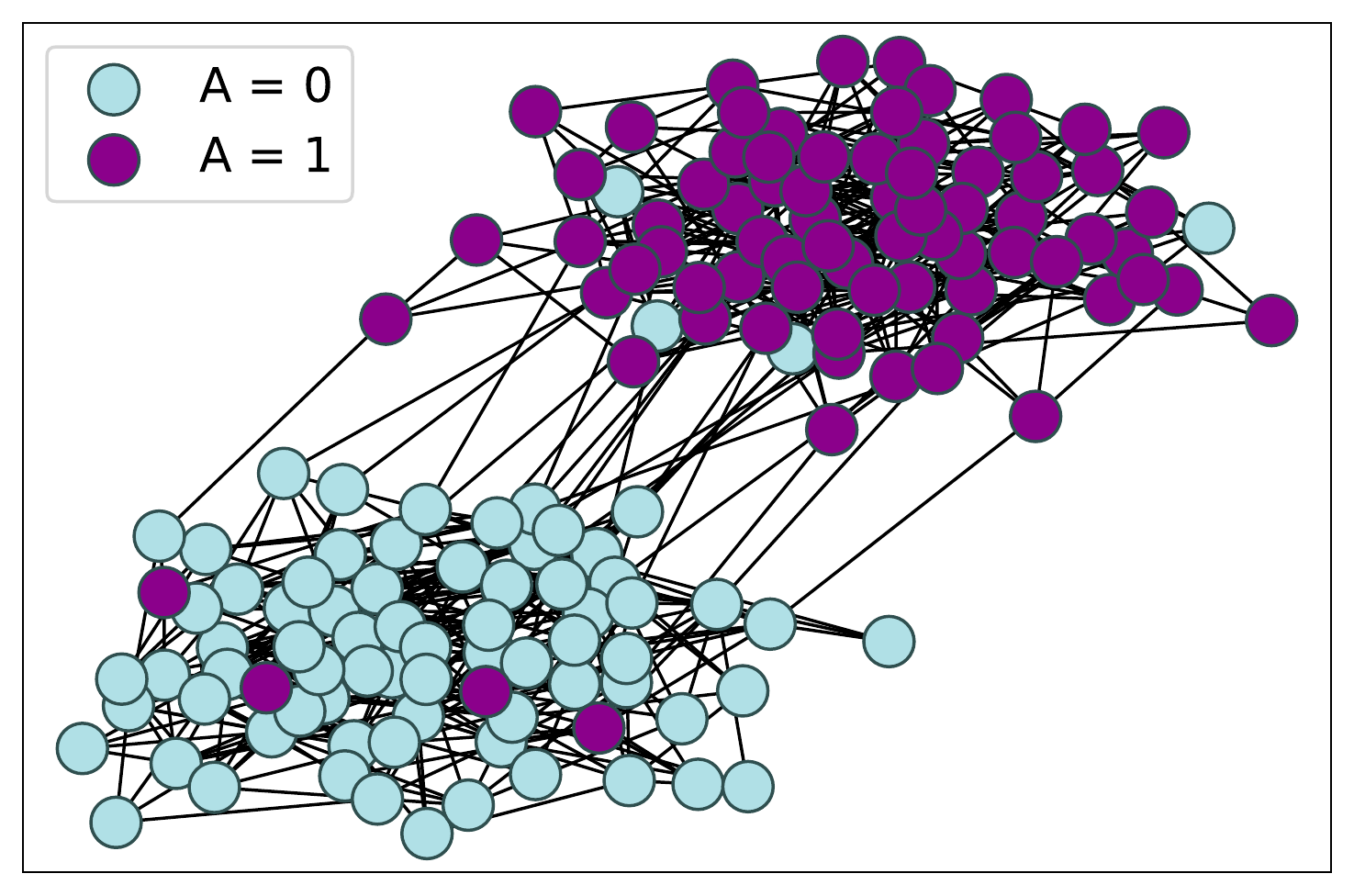}}
    \subfloat[\Gd]{\includegraphics[width=0.31\linewidth]{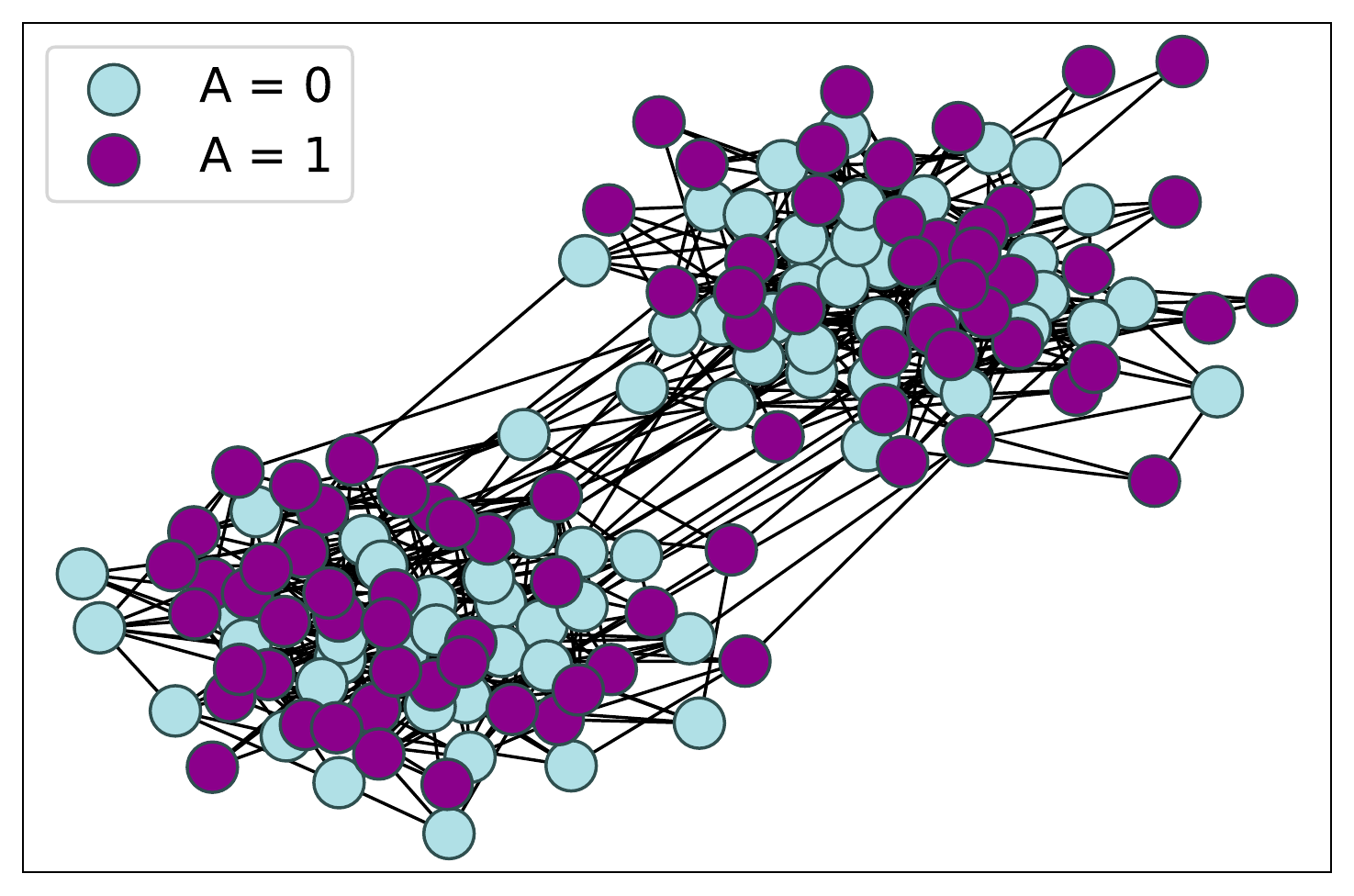}}
    \subfloat[\Gt]{\includegraphics[width=0.31\linewidth]{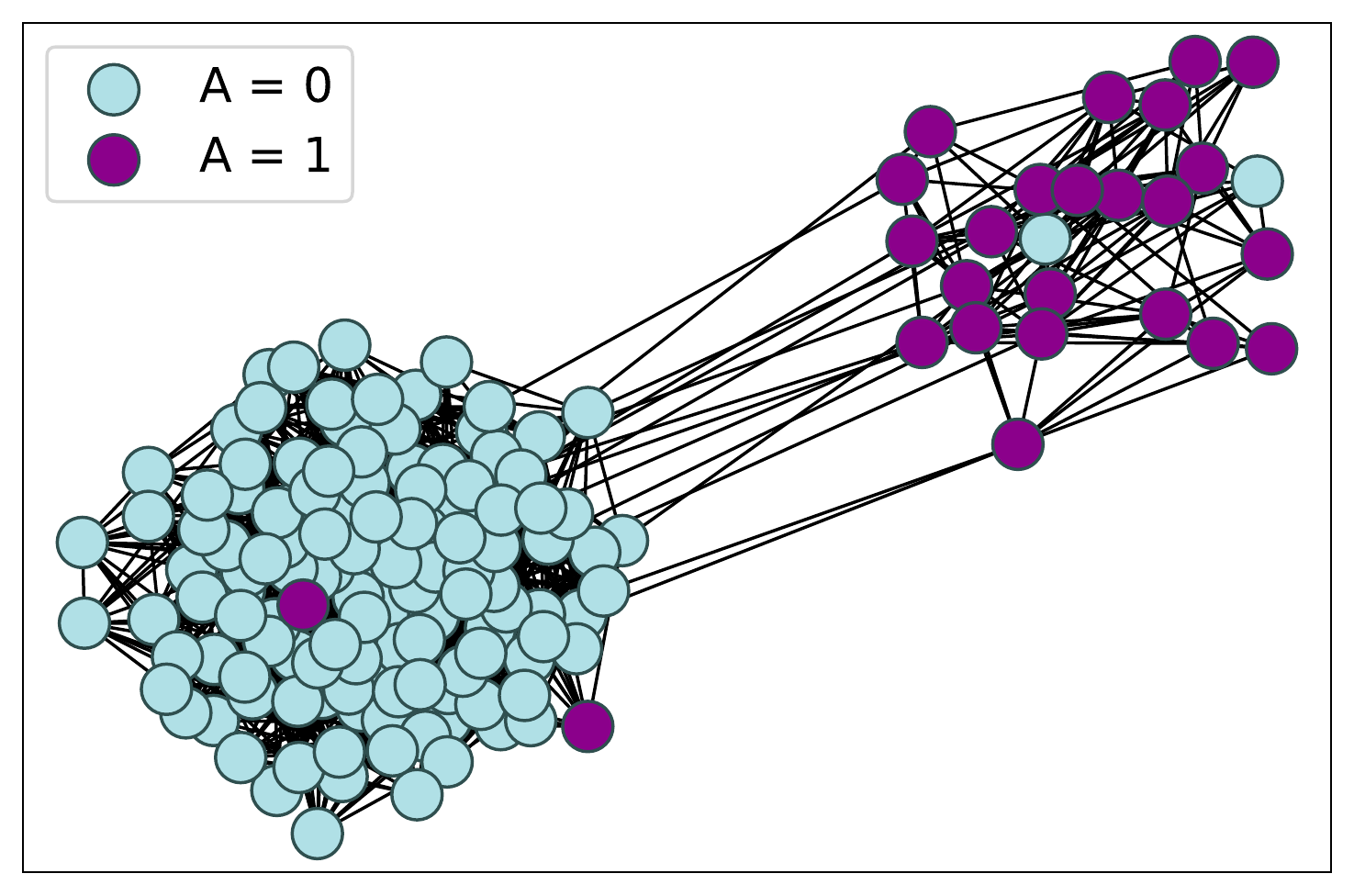}}\\
    \subfloat[\Gq]{\includegraphics[width=0.33\linewidth]{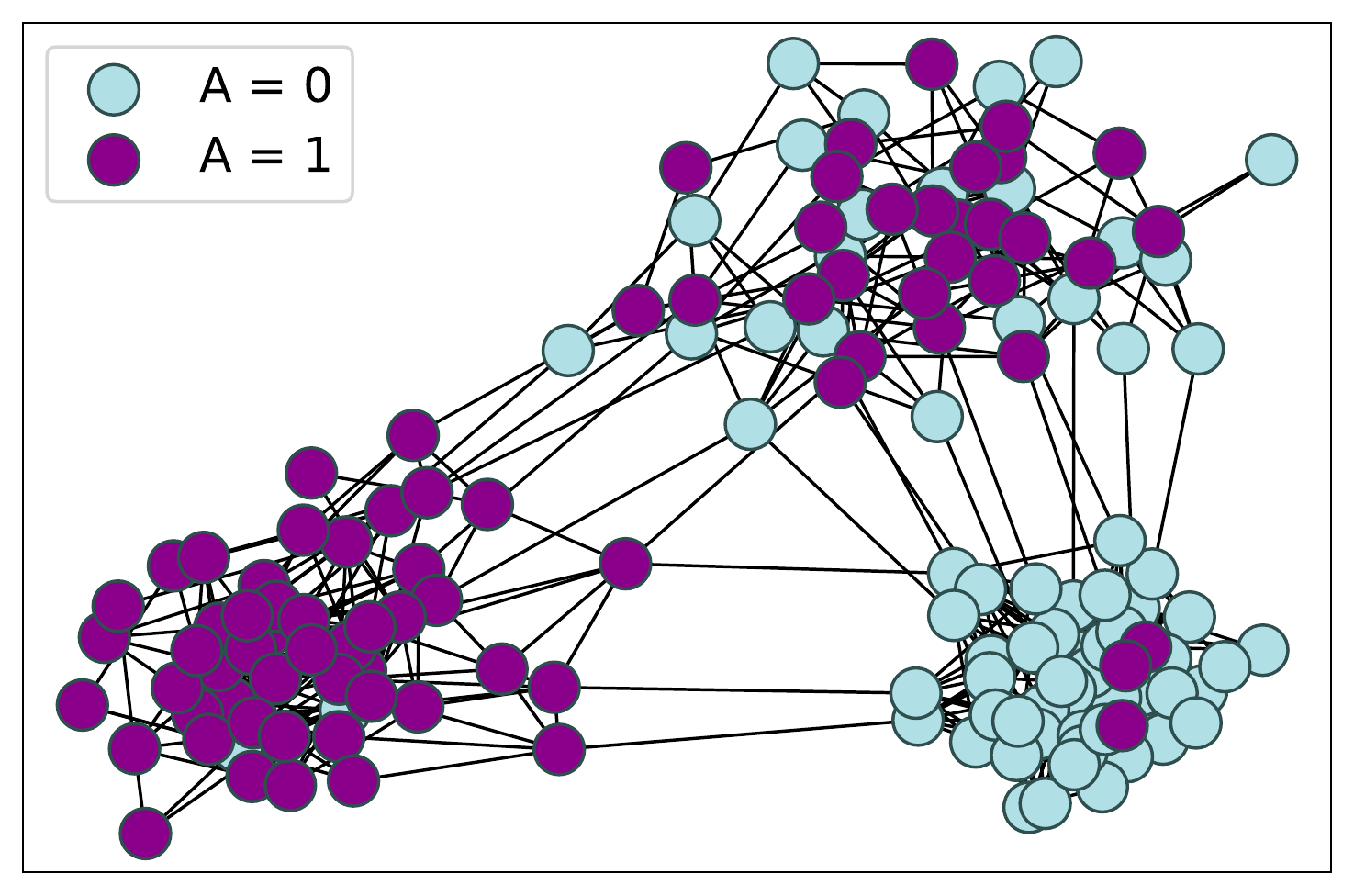}}
    \subfloat[\Gc]{\includegraphics[width=0.29\linewidth]{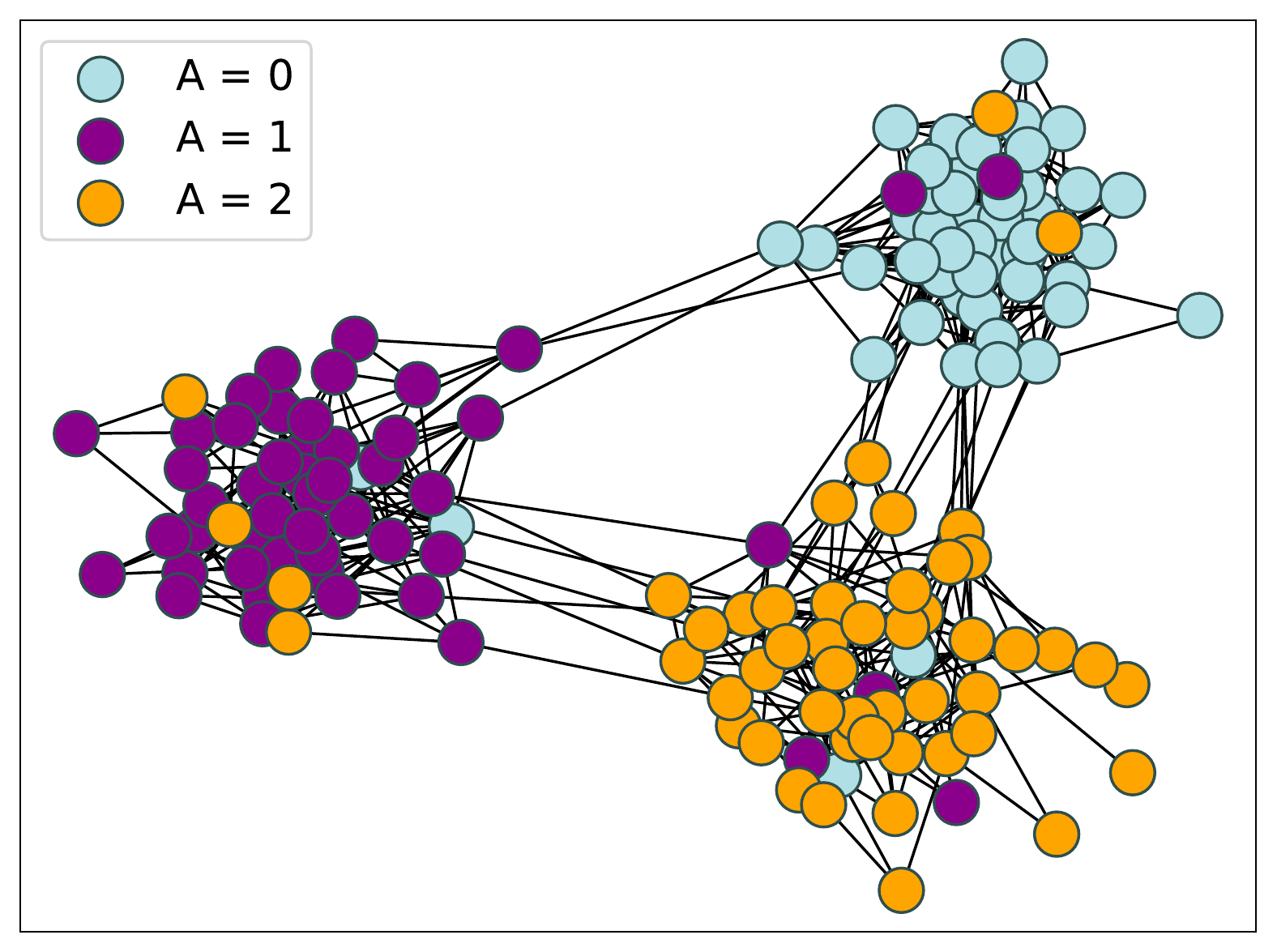}}
    \subfloat[\Gs]{\includegraphics[width=0.29\textwidth]{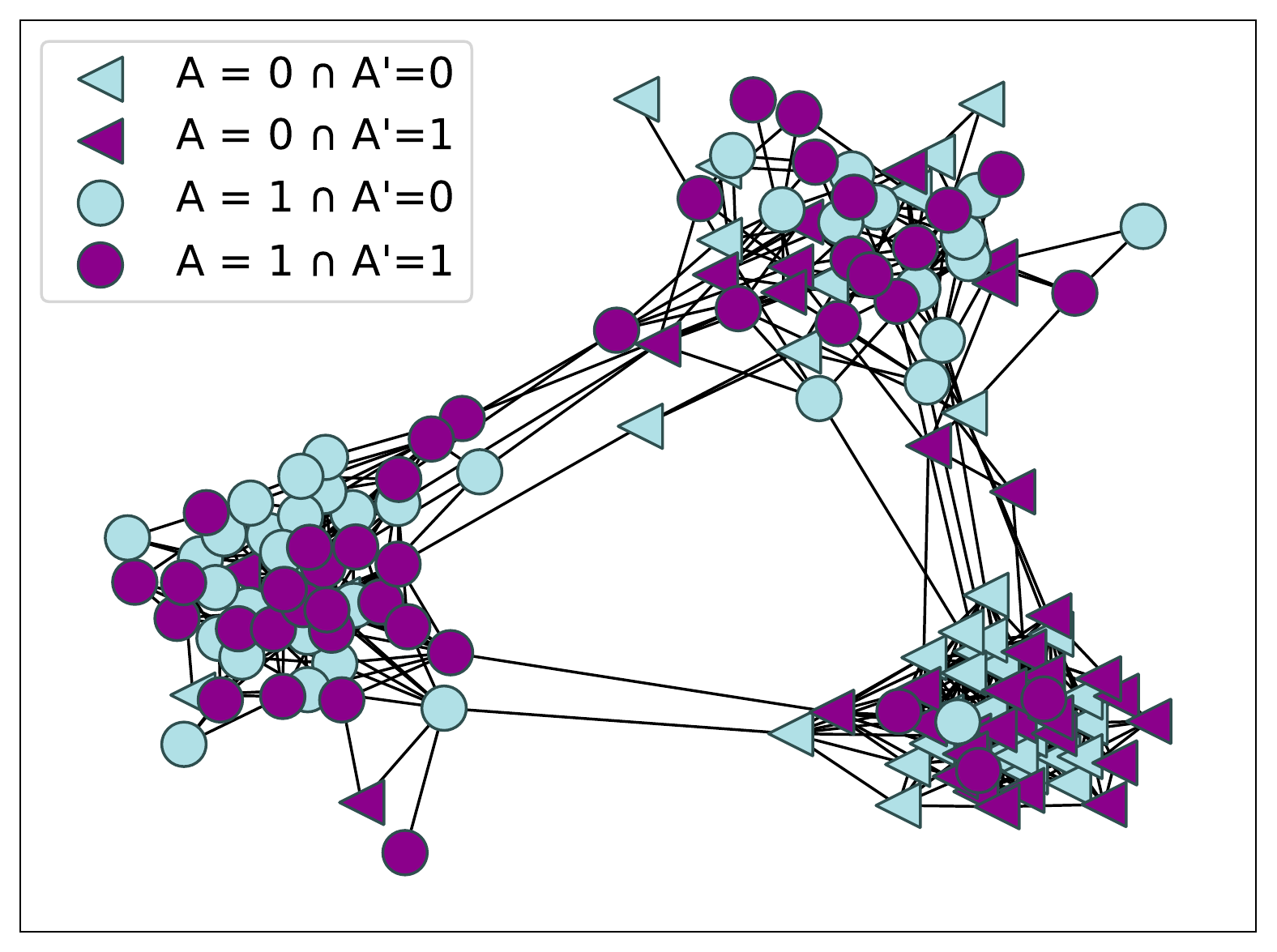}}\\
    \caption{Synthetic graphs proposed to address fairness in graphs. $\Gu-\Gc$ were initially proposed by \cite{laclau2020optimal}. $\Gs$ is a new scenario.}
    \label{fig:synthetic_graphs}
\end{figure}

Hereafter, we describe in more depth each of the graph structure:
\begin{itemize}
    \item[-] {\Gu} corresponds to a graph with two communities and a strong dependency between the community structure and the sensitive attribute $A$; 
    \item[-] {\Gd} has the same community structure as {\Gu}, but with $A$ being independent of the structure; 
    \item[-] {\Gt} corresponds to a graph with two imbalanced communities dependent on $A$ and a stronger intra-connection in the smaller community;
    \item[-] {\Gq} is a graph with three communities, two of them being dependent on $A$, and the third one being independent;
    \item[-] {\Gc} is a multiclass version of {\Gq} with three communities, each of them being dependent on one of the modalities of the protected attribute;
    \item[-] {\Gs} corresponds to the multiple attributes case, that only very few approaches can handle. The two protected attributes are distributed along with the three communities. The first attribute, denoted by $A$ is correlated with the community structure, while the second one, $A'$ is randomly distributed.
\end{itemize}  
Figure \ref{fig:synthetic_graphs} illustrates each scenario with a graph generated according to the parameters of this latter. 

\begin{table}[!htpb]
\caption{Parameters used for the graph generation. \textit{Cluster} implies that the value of $A$ is almost equal to the community identifier, while \textit{random} implies that $A$ is randomly generated and is independent of the community structure of the graph. $\bar{r}$ is the average of the assortativity coefficients obtained for 10 simulations with similar parameters,  \cris{$k$ is the number of modalities of $A$ and $|A|$  the number of protected attributes. Size indicates the size of the communities}}
\label{tab:parameters}
    \centering
      \resizebox{0.6\textwidth}{!}{
    \begin{tabular}{ccccccc}
    \toprule
    Graphs & $A$& $k$& Size & 
    |A| &$\bar{ r}$\\ 
    \midrule
     {\Gu}   & cluster &2& $\begin{pmatrix}75 &75 \end{pmatrix}$& 1&$.768\pm{.055}$\\
     {\Gd}   & random &2& $\begin{pmatrix}75 &75 \end{pmatrix}$& 1& $.001\pm{.035}$\\
     {\Gt}  & cluster &2& $\begin{pmatrix}125 &25 \end{pmatrix}$&1&$.711\pm{.082}$\\
     {\Gq}   & random \& cluster & 2&$\begin{pmatrix}50 &50 &50 \end{pmatrix}$& 1&$.576\pm{.037}$\\
     {\Gc}   & cluster &3& $\begin{pmatrix}50 &50 &50 \end{pmatrix}$& 1& $.732\pm{.041}$ \\
     {\Gs}  & cluster &2&$\begin{pmatrix}50 &50 &50 \end{pmatrix}$ & 2& -\\
    \bottomrule
    \end{tabular}
    }
\end{table}

\subsection{Real-world Graphs}

Hereafter, we propose a list of the real-world graphs commonly used in papers dedicated to fairness for graphs. Table \ref{tab:graph_stat} provides both statistics on the graphs and information related to the protected attribute(s). This list along with a link to download the benchmarks is available on the repository of the survey.

\smallskip

\subsubsection{Benchmarks with binary sensitive attribute}
~~

\polblogs \cite{adamic2005political} is a political blogosphere network of the US for February 2005. In this graph, the nodes represent blogs and vertices represent the hyperlinks between the two blogs. In this context, the political leaning associated with the node, which is either liberal (0) or conservative (1) is the sensitive attribute. 

\smallskip

\fbe~  and \google~ \cite{mcauley2012learning} are two collections of ego networks of the Facebook app and Google+ users, respectively. For \fbe, the combined version consists of the aggregated networks of Facebook friend lists for ten individuals. \google~ comprises ego-networks of 133 users whose information was publicly accessible at the time of the crawl and included only the users who had shared at least two circles. In both cases, the sensitive attribute is the gender information corresponding to each node. 
A particularity of \fbe~ is that its structure is not particularly dependent on the sensitive attribute, making this latter a good candidate for a sanity check (i.e., one should expect stable results with or without group fairness constraints).

\smallskip

\nba~ \cite{enyan2021} contains performance statistics of players in the 2016-2017 season along with information such as nationality, age and salary for 400 NBA basketball players. The graph connecting these NBA players is obtained using their relationships on Twitter. All the information can be considered a potentially sensitive attributes: the nationality of the players which is binary (American players vs. overseas players) but also their age and salary.

\smallskip

\subsubsection{Benchmarks with categorical sensitive attribute}
~~

\lastfm~ Asia Social Network \cite{feather} is a network collected from the \lastfm~ public API in March 2020. In this network, the nodes represent the \lastfm~ users and the edges represent the mutual follower relationships between the users. The sensitive attribute is the country of the user and takes 18 modalities. One should note that a particularity of the graph is that the distribution of the modalities of the country is very imbalanced.

\smallskip

\fbp~ \cite{rozemberczki2019multiscale} is a page-page graph of verified facebook sites. The nodes correspond to the official Facebook pages and there exists a link between the nodes if there exist mutual likes between the sites. The page-type is treated as the sensitive attribute. The page-type can take 4 values: \textit{company}, \textit{government}, \textit{politician} and \textit{tvshow}.

\smallskip

\pokec~ \cite{takac2012data} is a very popular widely used online social network in Slovakia (similar to Facebook and Twitter). Nowadays, \pokec~ counts approximately 1.6 million users. This dataset comprises anonymized data of the whole network. Since the dataset is large, in \cite{enyan2021}, the authors sample two datasets (based on the provinces the users belong to) from the original \pokec~ dataset namely \pokec-z and \pokec~-n. Several attributes can be used as sensitive including the region, gender and age of users.

\smallskip

\dblp~ is a co-authorship network constructed from the computer science bibliography database. The version proposed by \cite{buyl2020} is considered where the publications in specific conferences are taken into account. The continent is the sensitive attribute and it is extracted from the author's country of affiliation. The continents Africa and Antarctica are not considered due to a strong under-representation in the data.

\smallskip
\begin{table}[!htpb]
    \centering
        \caption{Descriptive Statistics of Benchmark Graphs used in the context of Fair Edge Prediction. Attributes are categorized as Cat. (categorical) or Cont. (continuous). A ``--" indicates that the property cannot be computed (for instance the $r$ mixing coefficient for continuous attributes).}
    \label{tab:graph_stat}
    \resizebox{\textwidth}{!}{
    \begin{tabular}{lcccccc}
    \toprule[0.1em]
    &\multicolumn{6}{c}{Networks}\\
    \cmidrule{2-7}
    Properties& PolBlogs&Facebook-E&Pokec-z&Pokec-n&LastFM&Facebook-P\\
      \midrule
    \# of nodes&$1,490$&$4,039$&$67,797$&$66,569$&$7,624$&$22,470$\\
    \# of edges&$19,090$&$88,234$&$617,958$&$517,047$&$27,806$&$171,002$\\
    Type of attributes&Cat.&Cat.&Cat.&Cat.&Cat.&Cat.\\
    Attribute Name&Party&Gender&Region&Region&Country&Page-type\\
    \# of inter-group edges&$1,575$&$41,875$&$30,519$&$24,293$&$3,507$&$19,590$\\
    \# of intra-group edges&$15,203$&$46,359$&$587,439$&$492,754$&$24,299$&$151,412$\\
    Mixing Coefficient $r$&$.81$&$.09$&$.87$&$.89$&$.86$&$.82$\\
    \bottomrule[0.1em]
    Properties&MovieLens-1M&MovieLens-100K& DBLP&Reddit&Google+&NBA\\
      \midrule 
    \# of nodes& \#users $6,040$,\#movies $3,900$ &$\#$users $943$,$\#$movies $1,682$ &$3,980$&$385,735$&$4,938$&$400$\\
    \# of edges&$1,000,209$&$100,000$&$6,965$&$366,797$&$547,923$&$10,621$\\
    Type of attributes&Cat.,Cont.&Cat.,Cont.&Cat.&Cat.&Cat.&Cat., Cont.\\
    Attribute Name&(Gender, Occupation, Age)&(Gender, Occupation, Age)&Continent&Subreddits&Gender&(Salary,Country,Age)\\
    \# of inter-group edges&$-$&$-$&$669$&$-$&$-$&$(2,935; 4,166; -)$\\
    \# of intra-group edges&$-$&$-$&$5,887$&$-$&$-$&$(7,686;6,455;-)$\\
    Mixing Coefficient $r$&$-$&$-$&$.84$&$-$&$-$&$(.015,.218, -)$\\
    \bottomrule[0.1em]
    \end{tabular}
    }
\end{table}
\subsubsection{Bipartite graphs benchmarks}
~~

\movielens~ is a collection of bipartite graphs consisting of user-movie ratings, on a scale of one to five. Nodes can therefore be of two types (users or movies) and an edge indicates that a user has rated a particular movie. For each user, the data set provides meta-information 
including gender, occupation and age. The
user features are considered as sensitive attributes.  Two versions of this dataset are traditionally used to evaluate graph fairness contributions, namely \movielens-100K and \movielens-1M.

\smallskip

\reddit~ \cite{bose2019compositional} is a dataset from the popular discussion website where the registered members submit content. These posts are organized by subject covering various topics and called  "communities" or "subreddits". The dataset is described in \cite{bose2019compositional}. The authors propose considering the comments from the month of November 2017, and define an edge between a user and a community if the user commented at least once on that community during this period. The low-degree nodes were removed to obtain the final graph with 366K users, 18K communities, and 7M edges. Reddit does not have public user attributes and the obtained graph is bipartite, so the attributes needed to be assigned to the users as well as the communities. Certain subreddits / node communities were treated as \textit{sensitive nodes} and the sensitive attribute of the user was based on whether or not the user is connected to the sensitive community.

\section{Open Challenges}\label{sec:conclusion}
While fairness has gained significant attention in graph-based machine learning and mining over the past three years, much remains to be done to mitigate the problem of biased decisions for this particular type of data. In this section, we present some of the remaining challenges in fair graph learning, and an overview of opportunities for future research. Table \ref{tab:sota-properties} summarizes some of the main properties of the state-of-the-art models aforementioned to highlight some pitfalls or gaps of current existing contributions.


\begin{table}[ht]
    \centering
      \caption{Outline of related work in terms of fulfilled (\textcolor{ForestGreen}{\ding{51}}) and missing properties (\textcolor{red}{\ding{55}}) of the algorithms. Cat. stands for categorical while Cont. stands for continuous. $k$ denotes the number of modalities that $A$ can take and $|A|$ denotes the number of sensitive attributes. For the fairness criteria we distinguished between individual (Ind.), group (Group) and counterfactual (Count.) fairness.
 Note that this property indicates which criteria is the model based on to achieve fairness. Row color groups for methods falling in the same category. For \texttt{INFORM} and \texttt{REDRESS} the notion of sensitive attribute does not apply as it only considers individual fairness criteria.}
      \label{tab:sota-properties}
    \resizebox{0.9\textwidth}{!}{
    \begin{tabular}{@{}lccccccccccccc@{}}
      \toprule[0.1em]
     &  \multicolumn{2}{c}{Type of Graphs}&\phantom{}&\multicolumn{4}{c}{Attribute's properties}&\phantom{}&\multicolumn{4}{c}{Fairness Criteria}\\
    \cmidrule{2-3} \cmidrule{5-8} \cmidrule{10-13}
     \textit{Method}& Directed & Bipartite&&Cat.&$k>2$&Cont.&$|A|>1$&&Ind.&Group& Count.&\\
    \hline 
    \rowcolor{blue}
       \texttt{FairOT}&\yes&\no&&\yes&\yes&\no&\no&&\yes&\yes&\no&\\
    \rowcolor{blue}
	\texttt{MaxFair}&\no&\no&&\yes&\yes&\no&\no&&\no&\yes&\no&\\
       \rowcolor{blue}
       \texttt{FairDrop}&\no&\no&&\yes&\yes&\no&\no&&\no&\yes&\no&\\
       \rowcolor{cosmiclatte}
       \texttt{Fairwalk}&\no&\no&&\yes&\no&\no&\no&&\no&\yes&\no&\\
       \rowcolor{cosmiclatte}
       \texttt{Crosswalk}&\no&\no&&\yes&\yes&\no&\no&&\no&\yes&\no&\\
       \rowcolor{cosmiclatte}
       \texttt{DeBayes}  &\no&\yes&&\yes&\yes&\no&\no&&\no&\yes&\no&\\
       \rowcolor{cosmiclatte}
       \texttt{MONET}&\no&\no&&\yes&\yes&\no&\no&&\no&\no&\no&\\
       \rowcolor{cosmiclatte}
       \texttt{NIFTY}&\no&\no&&\yes&\no&\no&\no&&\no&\no&\yes&\\
       \rowcolor{cosmiclatte}
       \texttt{FairGNN}&\no&\no&&\yes&\no&\no&\no&&\no&\no&\no&\\
   \rowcolor{cosmiclatte}
       \texttt{FairVGNN}&\no&\no&&\yes&\yes&\no&\no&&\no&\yes&\no&\\
   \rowcolor{cosmiclatte}
	\texttt{DFGNN}&\no&\no&&\yes&\no&\no&\no&&\no&\yes&\no&\\
   \rowcolor{cosmiclatte}
	\texttt{FGNN}&\no&\no&&\yes&\no&\no&\no&&\no&\yes&\no&\\
   \rowcolor{cosmiclatte}
        \texttt{FairHIN}&\yes&\yes&&\yes&\no&\no&\no&&\no&\yes&\no&\\
   \rowcolor{cosmiclatte}
      \texttt{FairEGM}&\no&\no&&\yes&\yes&\no&\no&&\no&\yes&\no&\\
   \rowcolor{cosmiclatte}
      \texttt{FairAdj}&\no&\no&&\yes&\no&\no&\no&&\no&\yes&\no&\\
    \rowcolor{cosmiclatte}
        \texttt{UGE}&\no&\no&&\yes&\yes&\no&\no&&\no&\no&\no&\\
   \rowcolor{cosmiclatte}
	\texttt{HM-EIICT}&\no&\no&&\yes&\yes&\no&\no&&\no&\yes&\no&\\
   \rowcolor{cosmiclatte}
       \texttt{FIPR}&\yes&\yes&&\yes&\yes&\no&\no&&\no&\yes&\no&\\
    \texttt{InFoRM}&\no&\no&&--&--&--&--&&\yes&\no&\no&\\
\texttt{REFEREE}&\no&\no&&\yes&\no&\no&\no&&--&--&\no&\\
       \rowcolor{mistyrose}
       \texttt{CFC}&\yes&\yes&&\yes&\yes&\no&\yes&&\no&\no&\no&\\
       \rowcolor{mistyrose}
       \texttt{FLIP}&\no&\no&&\yes&\no&\no&\no&&\no&\yes&\no&\\
           \texttt{EDITS}&\no&\no&&\yes&\no&\no&\yes&&\no&\yes&\no&\\
       \texttt{REDRESS}&\no&\no&&--&--&--&--&&\yes&\no&\no&\\
        \texttt{FairInf}&\no&\no&&\yes&\yes&\no&\yes&&\no&\yes&\no&\\
	\texttt{FairCovering}&\yes&\no&&\yes&\yes&\no&\yes&&\no&\yes&\no&\\
	\texttt{FairTCIM}&\yes&\no&&\yes&\yes&\no&\yes&&\no&\yes&\no&\\
	\texttt{DivConstraint}&\yes&\no&&\yes&\yes&\no&\yes&&\no&\yes&\no&\\
	\texttt{AdvFairInfluence}&\no&\no&&\yes&\yes&\no&\yes&&\no&\yes&\no&\\
       \rowcolor{gainsboro}
       \bottomrule[0.1em]
    \end{tabular}
   }
    
\end{table}

\subsection{Non-iid Assumption and Homophily Effect}
The iid (independent and identically distributed) hypothesis is central in machine learning literature.  Thereby, most of the algorithms assume, implicitly or explicitly, that the elements belonging to the dataset are independently drawn from the same distribution. If this hypothesis seems likely most of the time, it does not hold anymore for graph data describing relationships between the elements.
Indeed, according to this hypothesis, the probability for node 1 of having N1 as a neighborhood should be independent from the probability for another node 2 to have N2 as a neighborhood; which is not true, notably in graphs with community structure. It is not without consequence on the graph-based probabilistic inference. For instance, it is commonly accepted that the joint distribution is factorizable into a product of marginal distributions but, in fact, it is an oversimplification, done to make the optimization tractable. Therefore, even if the empirical results suggest that this oversimplification does not hurt the model performance, it would be better to have models based on assumptions more in line with the relational nature of the data. To the best of our knowledge, the work presented in \citep{ma2021subgroup} is the first one that proposes an analysis of fairness for non-IID data. Note that this assumption is also generally overlooked in the definition of metrics proposed to evaluate dyadic fairness.

Another specificity of graphs that makes the fairness problem difficult to solve, relies on the homophily property of its topology. This property, frequently observed in practice, states that similar nodes, according to the sensitive attribute, are more likely to attach to each other than dissimilar ones. When it is not verified i.e. in the case of disassortativity, one can hope that the models provide results as good as those obtained without taking into account fairness constraints. A systematic evaluation of both situations, with and without homophily, should be encouraged.

\subsection{Fairness Measurements}

The recent increase of interest in the topic has not resulted in the literature in a consensual definition of what is fair. On the opposite, it has inspired a variety of metrics, most of them designed not to assess fairness but a specific type of bias as shown in Section \ref{sec:metrics}. If it is clear that fairness is a complex notion, difficult to grasp in a single way, it seems important to better understand the relationship between these various metrics and the underlying aspect of fairness that they aim to assess. 
As there is no model able to optimize the whole of these metrics, notably because some of them can be conflicting, a deeper comparative study would benefit not only the scientific community but also the practitioners who have to choose a model suited to their task and application. Moreover, it would also lead to a better understanding of to which extent these different metrics are compatible and to the design of models able to combine different fairness aspects or at least prioritize them. \cris{This point is particularly tricky in the case of graph data due on the one hand to the influence exerted by the nodes on their neighbors and on the other hand to the phenomena of assortativity or, on the contrary, of disassortativity.}

In the same way, the utility-fairness trade-off dilemma needs also be deepened. 
When the graph is homophilic, it seems harder to improve fairness without degrading model performance. For this reason, most of the works consider that there is a trade-off such that improving fairness reduces the model’s accuracy and vice versa. \cris{For instance, this is often the case for GNN-based models where the objective function is modified to incorporate a term that mitigates the bias or for influence maximization models that relies on a utility function integrating a fairness constraint.}
However, it has also been shown, under reasonable assumptions, that fairness and performance are not necessarily opposite, at least for specific tasks and bias types \cite{WickpT19}. Revisiting this trade-off by analyzing the conditions where fairness and performance are compatible is also an open question. 

\subsection{Multiple Sensitive Attributes or Missing Sensitive Attribute}

Another challenge concerns the sensitive attributes themselves. As indicated in Table  \ref{tab:sota-properties}, most works focused on only one attribute, most often categorical taking only two modalities while the case of numerical sensitive attributes should also be considered. Moreover, exploiting simultaneously different attributes needs also to be studied. Investigating intersectional fairness does not mean improving the fairness of the model attribute per attribute but rather accommodating several sensitive attributes simultaneously such as age and gender to be sure that the model remains fair even for subpopulations like young women for instance.

\smallskip

An additional difficult issue related to the sensitive attribute might be the lack of availability. Whereas most of the methods assume that the sensitive information is fully accessible this is not necessarily the case. There are two interesting application scenarios where this particular challenge occurs. On the one hand, it is quite common for traditional online social networks to have missing attributes, as some of the fields filled by users are not mandatory (eg. age). In this case, one will eventually end up having a matrix of attributes with nodes (users) missing some relevant information. On the other hand, the problem of missing features is probably going to occur even more often in the coming years due to the evolution of European legislation and due to legal restrictions imposed by the General Data Protection Regulation (GDPR) that requires personal data protection \cite{Malgieri2020}. Indeed, according to the Artificial Intelligence Act, users protected information (eg. gender, relationship status) could be collected by companies to prevent possible discriminatory bias but the choice will be left to the users. Some approaches have been proposed to tackle the problem of missing attribute imputation in graphs. For instance, in \cite{rossi2021}, the authors proposed a model based on feature propagation. Most of these approaches, however, rely on the assumption of a strong homophily of the graph and rely on it to efficiently impute missing values. The latter hypothesis, as explained throughout this survey can carry a threat of reinforcing the bias in the data if the latter contains it.
This emerging research line is thus of high practical importance. 

\subsection{Benchmark and Generator}
The lack of benchmark datasets is another limit nowadays for studying and comparing both the different metrics and the methods for debiasing the data. This is especially true for studying fairness with missing sensitive attributes where the availability of the ground truth is crucial.
In this respect, just as generators already exist for building graphs having properties specified by the user, it would be very useful for the scientific community to have such generators to automatically create biased relational datasets where the user could control the degree or the type of bias introduced into the data as well as the correlation between the attribute to predict and the sensitive attribute. In Section \ref{sec:data}, we extend a framework, initially introduced by \cite{laclau2020optimal} but generators for graphs or even better, attributed graphs with community structures \cite{Lancichinetti2008, Largeron17} or assortatively mixed networks \cite{Newman_2003} could also be a good basis for developing additional tools.
Of course, real benchmark datasets would be also very complementary, and companies that collect a large amount of data containing sensible information should be encouraged to diffuse their datasets after anonymization. In this way, they could also demonstrate that they effectively respect fairness conditions in their automatic decision-making processes.

\subsection{Heterogeneous Graphs}

Finally, we believe that while the number of contributions keeps increasing, methods dedicated to heterogeneous graphs are missing (see Table \ref{tab:sota-properties}). It is particularly true for methods able to handle directed graphs, since GNNs approaches are not able to differentiate between directed and undirected edges. Similarly, contributions for bipartite graphs are also quite scarce. The main challenge being that most fairness metrics and fair models consider and compare the attributes between two nodes. However, this is not always available as in some bipartite graphs, such as \textsc{MovieLens} for instance, the notion of sensitive attribute for a movie does not apply, while it is available for users. Finally, graphs with a strong imbalanced distribution of the protected attributes also represent a challenge, and most of the existing approaches have not been evaluated in such scenarios. Indeed, in sociology, several criteria are used to identify potential sources of segregation in personal networks, among which we can find the relative group size. This is a problem that was also identified in \citep{laclau2020optimal}.

In this survey, we aim to give an overview of fair graph machine learning models. We focused on fairness definitions and categorized existing contributions according to several criteria: the type of fairness and where they operate in the learning process. As for the open challenges pointed out, one should note that these do not constitute an exhaustive list. They just give a few future research directions and other ones, related to the more general topic of fairness in machine learning, could also be investigated. Due to the strong societal impact of equity issues today, we hope that this survey will motivate researchers to tackle these open questions in the near future.

\bibliographystyle{ACM-Reference-Format}
\bibliography{main_arxiv}

\appendix

\end{document}